\documentclass[lettersize,journal]{IEEEtran}
\usepackage{amsmath,amsfonts}
\usepackage{mathtools,amssymb}
\DeclareMathOperator*{\argmax}{argmax}
\DeclareMathOperator*{\argmin}{argmin}
\DeclareMathOperator*{\softmax}{softmax}
\DeclareMathOperator*{\topk}{top-k}

\DeclarePairedDelimiter\floor{\lfloor}{\rfloor}

\usepackage{bm}
\usepackage{algorithmic}
\usepackage{algorithm}
\usepackage{array}
\usepackage{xcolor}
\usepackage{enumitem}
\usepackage{colortbl}
\usepackage[caption=false,font=normalsize,labelfont=sf,textfont=sf]{subfig}
\usepackage{textcomp}
\usepackage{booktabs}
\usepackage{stfloats}
\usepackage{url}
\usepackage{verbatim}
\usepackage{graphicx}
\usepackage{cite}
\usepackage{scalerel}
\usepackage{tikz}
\usetikzlibrary{svg.path}

\definecolor{orcidlogocol}{HTML}{A6CE39}
\tikzset{
  orcidlogo/.pic={
    \fill[orcidlogocol] svg{M256,128c0,70.7-57.3,128-128,128C57.3,256,0,198.7,0,128C0,57.3,57.3,0,128,0C198.7,0,256,57.3,256,128z};
    \fill[white] svg{M86.3,186.2H70.9V79.1h15.4v48.4V186.2z}
                 svg{M108.9,79.1h41.6c39.6,0,57,28.3,57,53.6c0,27.5-21.5,53.6-56.8,53.6h-41.8V79.1z M124.3,172.4h24.5c34.9,0,42.9-26.5,42.9-39.7c0-21.5-13.7-39.7-43.7-39.7h-23.7V172.4z}
                 svg{M88.7,56.8c0,5.5-4.5,10.1-10.1,10.1c-5.6,0-10.1-4.6-10.1-10.1c0-5.6,4.5-10.1,10.1-10.1C84.2,46.7,88.7,51.3,88.7,56.8z};
  }
}

\newcommand\orcidicon[1]{\href{https://orcid.org/#1}{\mbox{\scalerel*{
\begin{tikzpicture}[yscale=-1,transform shape]
\pic{orcidlogo};
\end{tikzpicture}
}{|}}}}

\usepackage{hyperref} 

\begin{document}

\title{When Less Is More: A Sparse Facial Motion Structure For Listening Motion Learning}

\author{T.T.~Nguyen~Nguyen \orcidicon{0000-0002-1543-3138},~\IEEEmembership{Student Member,~IEEE,}
        Q.~Tien~Dam \orcidicon{0000-0003-2670-7314},
        D.~Tuan~Tran \orcidicon{0000-0001-7443-9102},~\IEEEmembership{Member,~IEEE,}
        and~Joo-Ho~Lee \orcidicon{0000-0003-1015-5615},~\IEEEmembership{Senior~Member,~IEEE}
\thanks{This paper was produced by the IEEE Publication Technology Group. They are in Piscataway, NJ.}
\thanks{Manuscript submitted October 10, 2024;}
\thanks{$^{1}$ T.T. Nguyen Nguyen and Q.Tien Dam are with the Graduate School of Information Science and Engineering, Ritsumeikan University, Osaka, Japan.}
\thanks{$^{2}$ Joo-Ho Lee is with the College of Information Science and Engineering, Ritsumeikan University, Osaka, Japan.}
\thanks{$^{2}$ D. Tuan Tran is with the Faculty of Data Science, Shiga University, Japan.}
}

\markboth{Journal of \LaTeX\ Class Files,~Vol.~14, No.~8, August~2021}%
{Shell \MakeLowercase{\textit{et al.}}: A Sample Article Using IEEEtran.cls for IEEE Journals}

\IEEEpubid{0000--0000/00\$00.00~\copyright~2021 IEEE}

\maketitle

\begin{abstract}
Effective human behavior modeling is critical for successful human-robot interaction. Current state-of-the-art approaches for predicting listening head behavior during dyadic conversations employ continuous-to-discrete representations, where continuous facial motion sequence is converted into discrete latent tokens. However, non-verbal facial motion presents unique challenges owing to its temporal variance and multi-modal nature. State-of-the-art discrete motion token representation struggles to capture underlying non-verbal facial patterns making training the listening head inefficient with low-fidelity generated motion. This study proposes a novel method for representing and predicting non-verbal facial motion by encoding long sequences into a sparse sequence of keyframes and transition frames. By identifying crucial motion steps and interpolating intermediate frames, our method preserves the temporal structure of motion while enhancing instance-wise diversity during the learning process. Additionally, we apply this novel sparse representation to the task of listening head prediction, demonstrating its contribution to improving the explanation of facial motion patterns.
\end{abstract}


\section{\label{sec:introduction}Introduction}

\begin{figure*}[t!]
\centering
\includegraphics[width=1\textwidth, trim={0cm 4.5cm 0cm 0cm}, clip]{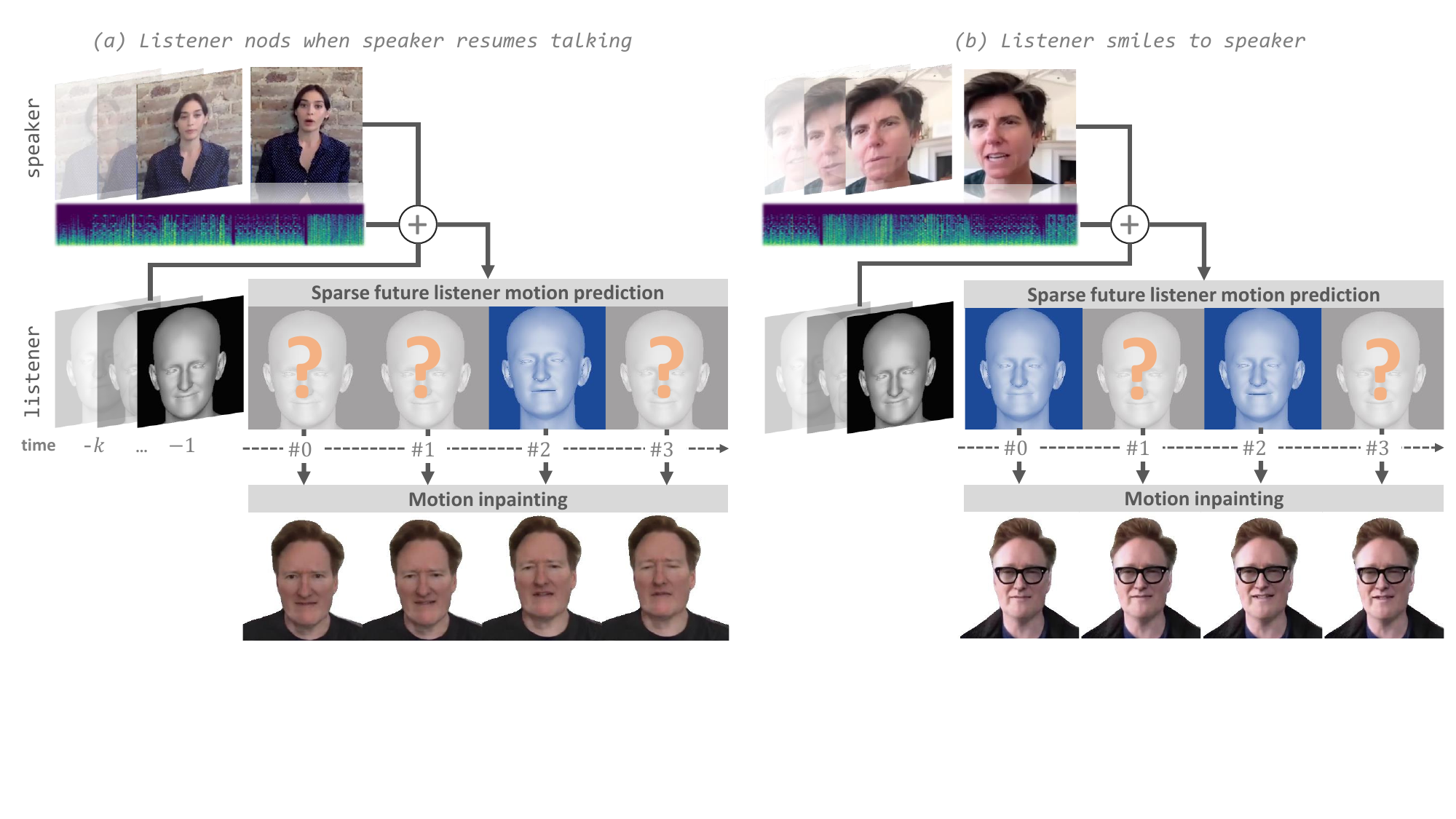}
\caption{\label{fig:intro_fig} \textbf{Listening head prediction with sparse token overview}. Our sparse representation captures key time steps from the listener's facial motion sequence, encoding temporal scale-varied and compound non-verbal facial motions, which generalize more effectively to the listening prediction task. The predictor determines the target facial motion to produce for future time steps and coordinates the transition of incoming frames toward specified facial expressions. By modeling both transition and key motion states as discrete tokens, this approach combines the robustness, stability, and flexibility of both discrete and continuous generative modeling.}
\end{figure*}

\IEEEPARstart{E}{ffective} non-verbal facial communication in the Human-Robot interaction has remained a highly sought-after topic \cite{vinciarelli2009social} over the last decade because of its importance in understanding the influence of human emotion dynamics in social interaction and the increasing potential applications in social-robot communication interface. A substantial body of literature within classical and modern psychology emphasizes the irreplaceable role of this behavior in conversation. Due to this demand, the task of predicting listening head movements has attracted renewed attention in recent years, with numerous studies and competitions focused on improving the generation of realistic non-verbal facial reactions conditioned on the multi-modal conversational context of two individuals. The high frequency, continuity, and compound nature of human facial patterns, which consist of multiple underlying action unit events, present significant theoretical and methodological challenges \cite{de2011facial}. {Current state-of-the-art Transformer-based techniques, which depend on continuous motion tokenization and next-token prediction, are hampered by the lack of clear boundaries} between prediction segments and the variability among similar motion patterns. Consequently, the generated motion tends to exhibit jittery transitions, abruptly shifting from one discrete pattern to another, derailing from the realism of the intended output. {Grouping nearby frames before tokenization or smoothing as post-process might mitigate the non-continuity problem, but not without the cost of diversity. On the contrary,} our study proposes a novel approach to learning a dynamic, continuous, and facial motion-friendly sparse representation, where keyframes are identified through an analysis-by-synthesis process. The proposed representation is then applied to the listening head prediction task and {and potentially other tasks, including micro-expression recognition, emotion recognition, and face verification. Our proposed unsupervised sparse facial expression model aligns with the apex-onset-offset framework in micro-expression and emotion recognition \cite{li2022deep, wang2024htnet}, where precise apex frame spotting is crucial. By effectively identifying keyframes in high-dimensional facial sequences, our approach enhances motion reconstruction. For face verification, filtering out low-informative and highly similar frames through sparse reconstruction reduces noise, leading to more stable verification in video data \cite{huber2024efficient}.} Overall, our main content focuses on three key areas:
\IEEEpubidadjcol
\begin{enumerate}
    \item Identifying keyframes from a facial motion sequence.
    \item Learning a novel keyframe-based sparse representation.
    \item Predicting facial feedback using the above sparse tokens.
\end{enumerate}

Although the listener's facial motion sequences are continuous by nature, recent advancements in speech, natural language processing, and human motion synthesis \cite{baevski2020wav2vec, li2021ai, zhu2023human} have led to the adoption of discrete motion tokens \cite{ng2022learning, dam2024finite, liu2024one}, rather than continuous representations \cite{jonell2020let, feng2017learn2smile}, for capturing human facial expression behavior. Techniques such as the Vector Quantized-Variational AutoEncoder (VQ-VAE)\cite{van2017neural} and Finite State Quantization (FSQ) \cite{mentzer2023finite} enable the reliable generation of nonverbal listening cues that generalize across novel contexts. Discrete representation-based approaches struggle with non-continuity, temporal variability, and latent space entanglement issues. While both human facial expressions and speech exhibit continuity in their nature, facial expressions are characterized by greater variability across individuals and instances. Additionally, facial motion lacks a well-defined system of supervised labels, complicating their analysis and interpretation.

To address these limitations, we propose a novel method called \textbf{Sparse Facial Motion Structure (SFMS)}, as illustrated in Figure \ref{fig:intro_fig}, which models continuous three-dimensional morphable model (3DMM) facial motion sequences as two types of discrete tokens: keyframes and transition frames. Keyframes are learned tokens that are discretized facial expressions at a given time step while transition frames are encoded by surrounding keyframes and their relative positioning to the keyframes. This design enables a sparse sequence of keyframe tokens to be decoded into high-fidelity, continuous facial motion, while preserving an efficient discrete latent space compatible with token-based predictors, such as Transformer models. The keyframe-based approach also aligns with the psychological theory that a universal facial communication system exists, as suggested by \cite{cuceloglu1972facial, kunz2023brain}, and is realized through the activation of various facial action units \cite{schmidt2001dynamics, cohn2007observer}. This theory states that facial motion can be divided into short segments dictated by a peak state along with inset and offset phases \cite{cohn2007observer}. However, such an approach has never been tested on non-verbal facial motions because of the lack of labeled data for keyframe signal guidance and effective strategies for isolating inset-offset phases in extended facial motion sequences. {In our study, our proposed sparse representation provides four-fold benefits:
\begin{enumerate}
    \item Firstly, for facial motion reconstruction on discrete latent space, the keyframe identification makes the quantizer more expressive and diverse, resulting in better reconstruction accuracy via a small codebook.
    \item Secondly, our sparse representation is the first implementation that can capture key expression changes aligned with human facial motor organization, where facial responses are driven by sparse, action-specific neural mechanisms \cite{kern2019human}.
    \item A next-token predictor powered by our sparse representation to generate listening head motion in dyadic conversations in a similar sparse manner.
    \item A robust evaluation framework based on two established datasets: Learning2Listen \cite{ng2022learning} and REACT23\cite{song2023react2023}, bridging previously separate lines of research.
\end{enumerate}}

\section{\label{sec:related_work}Related Work}

\subsection{Facial representation}

Facial representation modeling involves a diverse range of techniques. Early methods modeled facial motion using either two-dimensional (2D) facial landmarks \cite{feng2017learn2smile,huang2018generating, huang2017dyadgan} or a Facial Action Unit System (FACS) \cite{huang2018generating}. However, the advancement of facial representation techniques was initially hindered by information loss and limited dataset availability. This trend prompts a shift toward 3D approaches, particularly those that utilize 3DMM \cite{feng2021learning, wang2022faceverse} {where 3D mesh parameters are encoded into disentangled coefficient spaces such as identity, expression, pose, etc. For facial expression synthesis tasks such as listening head generation, extracting implicit temporal patterns of interaction between speech and facial while maintaining temporal consistency and naturalness is critical and challenging for typical dense facial representations where every video frame is processed despite their high similarity. This redundancy biases the training toward overfit trivial patterns while leaving challenging facial expressions underrepresented \cite{razavi2019generating}, especially in vector quantization-based techniques with finite codebooks \cite{van2017neural}. Noisy similar expressions also hinder the attention mechanism's effectiveness in the autoregressive model where short bursts of subtle motions are lost between long neutral facial motions. This work explores a new sparsity-emphasized representation where facial motion's key states are located and the in-between states are inpainted according to temporal relationship to nearby keyframes. Our hypothesis is by introducing sparsity to facial expression encoding, we lessen the burden of excavating the dynamic facial behavior for the generative modeling task.}

\subsection{Non-deep learning-based methods}

To model and predict a listeners facial behavior, various approaches have been broadly categorized into classical and deep learning-based techniques. Early methods employed classical machine learning and rule-based algorithms, including hybrid methods. Popular tools in this category include sparse 2D facial landmarks \cite{lee2020fuzzy}, emotion spaces\cite{lee2020fuzzy, gotoh2005face}, and dense point sets\cite{lee2014analysis}. Capturing motion dynamics relies on empirical kernel maps \cite{lee2014analysis}, linear subspaces \cite{chai2003vision}, and fuzzy systems \cite{lee2020fuzzy}. Although these techniques offer simplicity and interpretability, they are limited in terms of their diversity and flexibility. They require smaller datasets but restrict learnable motion groups to fixed categories, thereby limiting the range of potential motions.

\subsection{Deep learning-based methods}

Deep learning solutions for non-verbal facial motion modeling leverage data-intensive architectures to learn latent subspaces and generate listener head motions based on conversational context. These data-driven approaches significantly enhance scalability and generalization across diverse facial motion patterns. Recent studies \cite{ng2022learning, song2023emotional, liang2023unifarn, jonell2020let, yu2023leveraging} primarily adopt either continuous or discrete {facial motion} representation frameworks. 

Continuous methods utilize advanced generative architectures such as variational autoencoders (VAEs) \cite{song2023react2023}, normalizing flows \cite{jonell2020let}, diffusion models \cite{yu2023leveraging}, and generative adversarial networks (GANs) \cite{huang2017dyadgan, huang2018generating}. These models effectively capture complex motion variations but suffer from high computational costs and instability due to mode collapse \cite{ng2022learning}. 

{Hybrid approaches incorporate simpler deep learning architectures, including shallow perception \cite{gotoh2005face}, recurrent neural networks (RNNs) \cite{matsui2010model}, and long short-term memory (LSTM) networks \cite{dermouche2019generative, zhou2022responsive}, or recently Diffusion-based model \cite{liu2023mfr, yu2023leveraging} which reduce manual decision-making but remain constrained by the limitations of continuous latent spaces.} While continuous representations generate smoother and more expressive facial motions \cite{jonell2020let}, they are computationally expensive and difficult to control. Conversely, discrete methods offer improved stability and lower training costs \cite{montero2022lost}, yet they often struggle with motion fidelity and continuity constraints. {Continuous latent space methods as mentioned before, are straightforward, allowing more fine control and precision, but typically suffer from dull generation, error accumulation (feedback drift)\cite{liu2024listenformer}. Transformer with non-autoregressive decoding and diffusion models (to inject variability) showed to mitigate these issues, at the cost of higher computation \cite{liu2024customlistener, liu2023mfr}.}

Discrete representation approaches encode facial motion into a symbolic latent space, capturing expression variations using discrete tokens that aim to reconstruct the original motion while preserving key expression characteristics. Prior studies have shown that discrete methods can outperform continuous ones, particularly in low-resource settings \cite{ng2022learning, montero2022lost}. This is typically achieved via vector quantization \cite{van2017neural}, which enables token selection to be optimized through teacher-forcing strategies guided by speaker input \cite{ng2022learning} or affective cues \cite{song2023emotional, song2023react2023, liang2023unifarn}.

{However, discrete latent spaces face two key limitations. First, vector quantization introduces information loss, often causing jerky and discontinuous motions. Rare but meaningful motion patterns may be merged with more common ones into a single token, reducing output diversity and accuracy \cite{liu2022reduce, lazebnik2008supervised, song2023emotional}. Second, ensuring smooth transitions between discrete tokens is particularly challenging for subtle expressions and micro-expressions. Several methods \cite{song2023emotional, lazebnik2008supervised, song2023react2023, liang2023unifarn} addressed these issues by either assigning multiple tokens per timestep or incorporating additional modalities such as emotion label; L2L \cite{ng2022learning} encodes multiple frames into a single token; FSQ \cite{dam2024finite} leverages a large, efficient codebook with Transformer-based modeling to mitigate information loss. While emotional cues can improve expressiveness, they are expensive to annotate and, therefore, do not scale well. Frame-wise and grouped tokenization each introduce trade-offs: frame-wise encoding captures high-frequency motion but can lead to instability, while grouped approaches smooth transitions at the cost of expressiveness and temporal precision.}

{To the best of our knowledge, existing non-verbal facial motion discretization methods encode all frames uniformly into a latent space—an approach we refer to as dense discretization. In contrast, our proposed sparse method identifies keyframes selectively, enabling a more adaptive continuous-to-discrete mapping. As information loss constraints \cite{lazebnik2008supervised, mentzer2023finite} remain a theoretical challenge, and high-quality auxiliary affective cues such as eye-blinks \cite{song2023emotional}, emotional classes \cite{song2023multiple}, and action units (AUs) \cite{song2023react2023} are expensive to upscale with existing non-verbal related dataset, the sparse keyframe semantic context that we proposed may serve as a new unsupervised instrument and improve the code-to-motion translation between discrete prediction codes. Our selective encoding reduces redundancy in token representation while preserving motion fidelity, ultimately achieving a more efficient and adaptable balance between expressivity and computational efficiency.}

\section{Preliminaries}
\label{sec:prelim}

{For listening head prediction tasks, several datasets have been introduced, each offering unique characteristics in terms of facial expression representation, subject diversity, and data availability per individual. Notable examples include MIMICRY \cite{sun2011multimodal}, VICO \cite{zhou2022responsive}, Learning2Listen (L2L) \cite{ng2022learning}, REACT23 \cite{song2023react2023}, and Realtalk \cite{geng2023affective}. This study employed two datasets for motion representation learning and listening head prediction tasks.}  

\begin{itemize}
    \item \textbf{Dataset 1}: L2L \cite{ng2022learning}, with DECA \cite{feng2021learning} - full head metrical reconstruction facial features. {L2L provides a large number of training datasets for four subjects (72 hours, 6 identities).}
    
    \item \textbf{Dataset 2:} REACT\cite{song2023react2023} with FaceVerseV2\cite{wang2022faceverse}{- a tight head non-metrical reconstruction, instead featuring facial expression extracted from a large number of subjects (71.8 hours, 159 identities).}
 \end{itemize}
 
 {Both datasets encompass distinct design choices. L2L's DECA coefficients encode full-face features learned by ResNet50 \cite{feng2021learning}, whereas FaceVerseV2\cite{wang2022faceverse} prioritizes non-metrical, tight-head reconstruction for lower-quality but real-time applications. In terms of orientation, L2L emphasizes personalized facial style reconstruction that captures a rich subject-specific facial behavior latent space, while REACT requires the model to generalize across a large population, making it a strong benchmark for robustness and generalization in the listening head prediction task.}

\begin{figure*}[ht!]
\centering
\label{fig:data_flow}
\includegraphics[width=1.0\textwidth, trim={0.3cm 1.5cm 1.9cm 1.8cm}, clip]{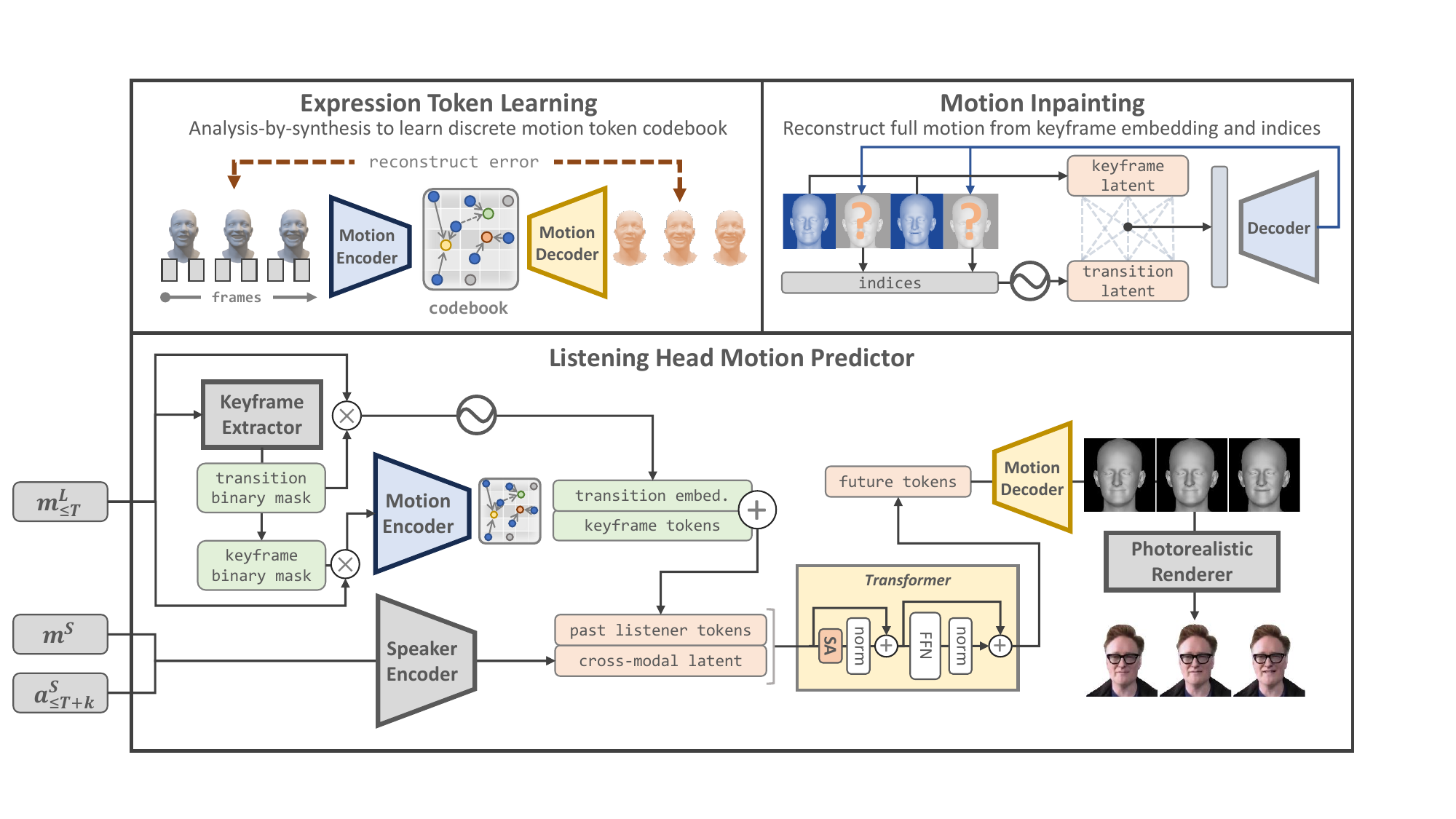}
\caption{\textbf{Training Pipeline Overview}. Our model learns to represent continuous motion as discrete tokens of key and transition frames with enhanced accuracy and fidelity. The proposal comprises two phases: reconstruction (top) and listening head motion prediction (bottom). The reconstruction task (top) includes two sub-modules: the expression token learning and motion {inpainting} models. The expression token learning model encodes a facial motion sequence into a finite set of discrete tokens, while the motion infilling model interpolates the blanks between these tokens with intermediate states. The prediction phase utilizes the trained reconstruction module to predict future facial tokens in a next-token prediction task, where the model must decide whether to react with a transition state or a key state that interrupts the current motion.}
\end{figure*}

\subsection{DECA and Learning2Listen}

\textbf{Facial feature}-wise, L2L \cite{ng2022learning} dataset proposes modeling listener facial motion as expression parameters $\psi\in\mathbb{R}^{50}$ and the pose codes $\theta\in\mathbb{R}^6$ (head pose $\mathbb{R}^3$ and jaw pose $\mathbb R^3$). L2L represents the listeners and speaker's facial animations using \textbf{DECA} \cite{feng2021learning} coarse features that regress a parametric face model based on FLAME \cite{li2017learning} geometry from a red, green, blue (RGB) image. DECA maps the subject identity $\delta \in\mathbb{R}^{128}$, expression $\psi\in\mathbb{R}^{50}$, and head pose $\theta\in\mathbb{R}^6$ features onto a 3D FLAME head mesh ($n=5023$ vertices) \cite{feng2021learning}. The mapping model $\mathcal{M}$ is defined as

\begin{equation}
    \mathcal{M}(\bm\delta,\bm\psi,\bm\theta) = \mathcal{W}(\textbf{T}, \textbf{J},\bm\theta,\bm{W})
\end{equation}

Facial expression transitions, represented by expression $\psi$ and pose $\theta$, are modeled by the blend skinning function $\mathcal{W}$, which rotates mesh vertices $\bm T \in \mathbb{R}^{3n}$ around joints $\bm J \in \mathbb{R}^{3k}$ and smooths them using blendweights $\bm W \in \mathbb{R}^{k \times n}$, $n$ denotes total number of head mesh model vertices.

\textbf{Speech feature}-wise, the speaker's audio is processed into $4T \times 128$ Mel-spectrogram features for every $T$ frames. The dataset, comprising 72 hours of 30-frames per second (FPS) video, focuses on interactions involving five program hosts. Although L2L provides diverse facial motion samples, each segment is limited to 64 frames and is imbalanced across subjects. Nevertheless, it offers a valuable collection of listener-speaker interactions in dyadic conversations.

\subsection{FaceVerseV2 and REACT}

The \textbf{REACT} dataset integrates data from the NoXI and RECOLA datasets, and comprises dyadic conversations conducted in an online conferencing format between interviewers and candidates. The training data includes 1,585 videos from the NoXI dataset, amounting to 14 h of footage, and nine videos from the RECOLA dataset. The test set consists of 553 videos from NoXI and nine from RECOLA, totaling 6.7 h. Most importantly, there is no subject overlap between the training and test sets \cite{song2023react2023}. Each sequence within the dataset is 750 frames long and recorded at a rate of 25 FPS.

\textbf{Facial features} in \textbf{REACT} consist of 58 dimensions, with expression parameters denoted as $\psi\in\mathbb{R}^{52}$ and pose parameters as $\theta\in\mathbb{R}^6$. Unlike the DECA full-head model, \textbf{FaceVerseV2} \cite{wang2022faceverse} models a tightly cropped facial region. \textbf{FaceVerseV2} employs a 3D base model $\mathcal M$ controlled by shape $\pmb s\in\mathbb R^{120}$, texture $\pmb p \in\mathbb R^{200}$, and pose parameters $\theta\in\mathbb{R}^6$ (translation $\mathbb R^3$ and rotation $\mathbb R^3$).
\begin{equation}
    \begin{aligned}
        \mathcal M (\bm S_{base}, &\bm T_{base}, \bm\theta)\\
        \textrm{with}\quad S_{base} = \bar{S} + \sum_{i=1}^{120}{s_i\alpha_i}&\quad T_{base} = \bar{T}+\sum_{i=1}^{200}{t_i\beta_i}
\end{aligned}
\label{eq:faceverse_model}
\end{equation}
In \eqref{eq:faceverse_model}, $\bar{S}$ and $\bar{T}$ represent the mean shape and texture. The principal components for shape and texture are denoted by $\alpha \in \mathbb{R}^{3n \times 120}$ and $\beta \in \mathbb{R}^{3n \times 200}$, where $n$ is the number of vertices. Shape parameters in FaceVerseV2 are projected into the expression subspace $\psi$ using the Apple ARKit 52 blendshapes: $S = S_{base} + \sum_{i=1}^{52} \psi_i \gamma_i$, where $\bm{\gamma} \in \mathbb{R}^{3n}$ defines 52 principal facial expressions. The feature vector $\psi \in \mathbb{R}^{52}$ represents blend weights for combining micro-expressions.

\textbf{Speech features} of REACT are in raw format. We process them into Mel-frequency Cepstral Coefficients (MFCC) and Wav2Vec 2.0 speech tokens \cite{baevski2020wav2vec}. Similar to a recent study \cite{dam2024finite}, we found speech-to-text token-based features to be more effective representations of listener facial feedback predictions using the REACT dataset.

\subsection{DECA and FaceVerseV2 comparison}

Unlike DECA, which focuses on anatomically accurate head modeling, FaceVerseV2 is optimized for lightweight facial expression representation. However, as a non-metrical model, FaceVerseV2 lacks intrinsic scale control, necessitating the normalization of 2D facial frames. Its inability to disentangle identity-specific facial geometries limits its capacity for reconstructing sequential facial motions and achieving photorealistic rendering. Consequently, FaceVerseV2 is more sensitive to pose variations and less effective at capturing individual-specific facial behaviors. Nonetheless, its linear blending of principal expressions facilitates a more semantically meaningful loss decomposition compared to applying a uniform norm loss across all dimensions.

\section{Methodology}

\begin{figure*}[t!]
\centering
\includegraphics[width=0.8\textwidth, trim={1cm 3.5cm 3cm 0.5cm}, clip]{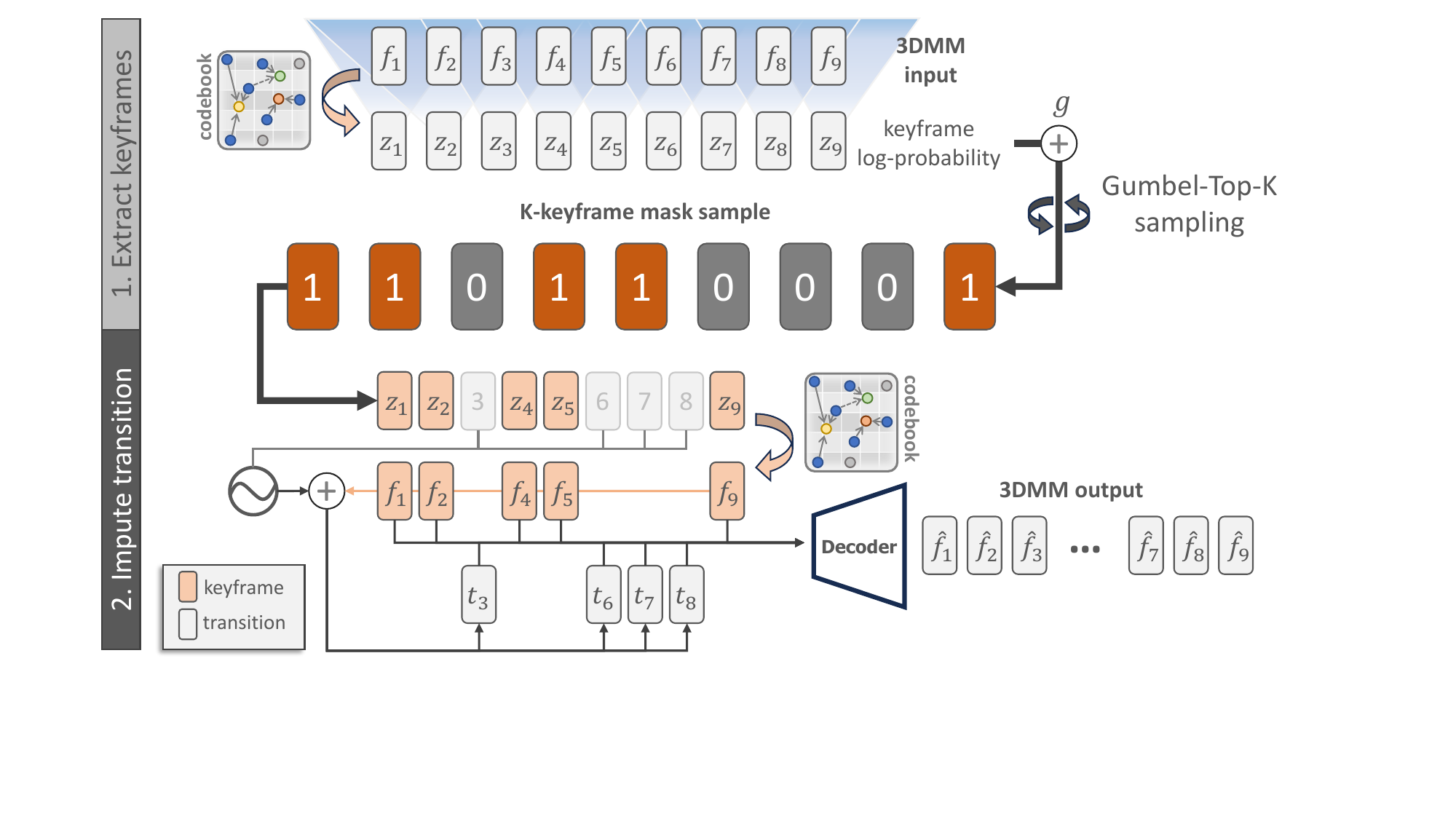}
\caption{\label{fig:masking} \textbf{Keyframe score learning and the reconstruction task}. The workflow starts by encoding 3DMM facial motion features into ranking logit scores to identify keyframes. Masks are sampled using the $\text{top-k}$ and Gumbel-Softmax functions for motion reconstruction. Keyframes are represented as vector quantized tokens, while transition frames are encoded positionally with keyframe information. The decoder reconstructs the original facial motion by combining keyframe and transition frame embeddings. Finally, the best reconstruction from the samples is used as the target for keyframe feature learning.}
\end{figure*}

\subsection{Data Processing}

\subsubsection{Facial Features}
Consider a dyadic conversation recorded over a discrete time horizon of $T$ frames, $\textbf{m}\in\mathbb{R}^{T\times d}$. From this point forward, facial expression sequences $\textbf{m}$ are universally referred to as 3DMM facial sequences, where the facial features in each frame are extracted using 3D face shape embedding, which can reconstruct the original facial expressions. This embedding is encoded using either DECA ($d=56$) or FaceVerseV2 ($d=58$) depending on the chosen dataset as introduced in Section \ref{sec:prelim}.

\subsubsection{Audio Features}
Two preprocessing pipelines are incorporated in our proposal:
\begin{enumerate}[label=(\alph*)]
    \item MFCC-based feature: These are widely used as audio representations. \cite{luo2023reactface, ng2022learning}.
    \item Wav2Vec 2.0\cite{baevski2020wav2vec}: This is well-known as a robust pretrained audio tokenizer for various speech-related tasks.
\end{enumerate}

Recent studies \cite{luo2023reactface, dam2024finite} have demonstrated that Wav2Vec 2.0 offers a significant performance gain over standard MFCC \cite{ng2022learning, song2023react2023} or the Geneva Minimalistic Acoustic Parameter Set (GeMAP) \cite{liang2023unifarn}. In this study, the Wav2Vec2-Base-960h variant was used for feature extraction from the REACT dataset. For the L2L dataset, we utilized the post-processed MFCC as the data maker, which does not include the raw audio data required for Wav2Vec encoding.\raggedbottom

\subsection{\label{sec:sparse} Sparse Facial Expression Representation}

In this section, we explore the representation of sequential 3DMM expression codes using a fixed number of vector-quantized keyframes interspersed with blank tokens or transition frames. The keyframes capture the peak expression moments, whereas the transition frames ensure locally dependent yet nuance-rich transitions between these peaks (Figure \ref{fig:compare_sparse}). Compared with dense representation approaches \cite{dam2024finite, ng2022learning}, the sparse representation method offers higher facial motion fidelity with the same number of expression tokens in the dictionary while easing the burden on expression token learning by reducing the overload on the finite codebook \cite{van2017neural} through a novel flexible inpainting strategy.

\subsubsection{Sparse Facial Motion Structure}

\begin{figure}[t!]
\centering
\includegraphics[width=0.5\textwidth, trim={4.5cm 0.8cm 5cm 0cm}, clip]{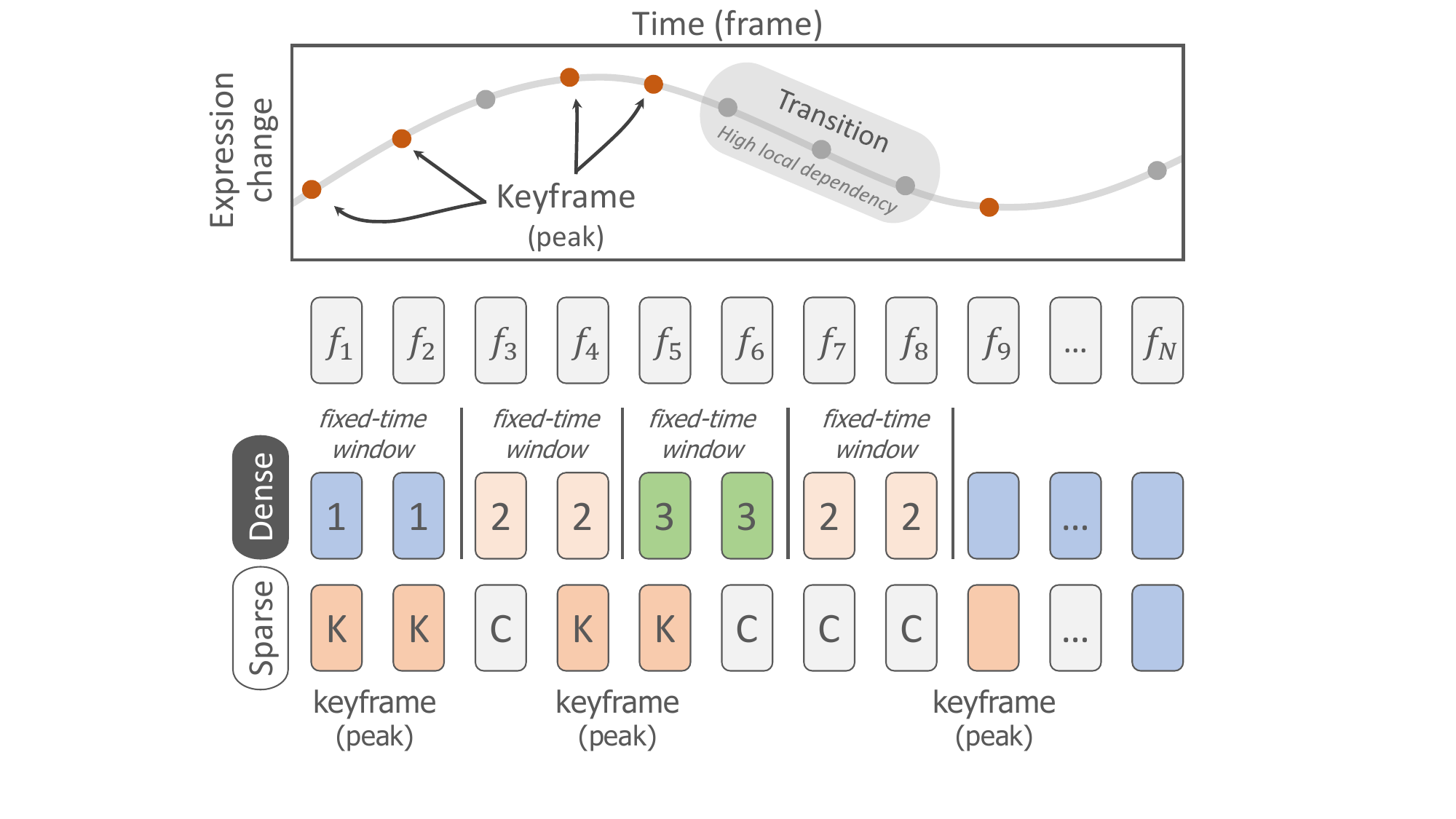}
\caption{\label{fig:compare_sparse}\textbf{Dense and Sparse Facial Motion Token Comparison}. In contrast to the dense structure, where 3DMM expression codes are uniformly encoded, our sparse structure selects keyframes and their positions to reconstruct locally dependent transition frames. This approach captures macro-level expression details while reducing computational complexity and minimizing information loss during the continuous-to-discrete conversion between keyframes.}
\end{figure}

In a continuous-to-discrete representation, a dense structure typically refers to the approaches in recent studies \cite{ng2022learning, liang2023unifarn, dam2024finite}, where each keyframe is uniformly encoded into a discrete token, similar to practices in domains such as speech, image, and body motion. Unlike linguistic units, facial expressions do not have well-defined boundaries, making it difficult for dense structures to effectively capture nuances, such as temporal variation, interruptions, or transitions between expressions. Our proposed sparse structure addresses this challenge by utilizing the local dependency of non-verbal facial motion to represent continuous facial expressions with sparsely distributed keyframes. The gaps between these keyframes are later filled using an imputation technique based on the expression states and positions; see Figure \ref{fig:compare_sparse}. Please note that, consistent with previous studies, our focus is on the temporal dynamics of coarse facial expressions, excluding fine details such as subtle skin wrinkles, which current state-of-the-art methods cannot reliably encode.

Formally, given a continuous $T$ frame-long facial motion sequence $\bm m = \{m_0,...,m_T\}\in\mathbb{R}^{T\times d}$, we define a binary mask $C(t)$ that classifies each frame position $t$ as either a keyframe or a transition frame; we aim to discretize this continuous sequence with sparsely distributed frame-wise discrete tokens from $N$-element dictionary $\mathcal D$. The $C(\cdot)$ function separates the $\bm m$ into $k$ key time steps $M$ and $(T-k)$ transition groups $G=\{G_k\lvert G_k=(m_{k+1},...,m_{k+K_i})\}$, which are approximated as $\hat M$ and $\Tilde G$, respectively in \eqref{eq:discrete_motion}.
\begin{equation}
    \begin{aligned}
\bm{m} \approx \hat{M}& = \hat M_0 \cup \bigcup^{K-1}_{k=1}{\bigg(\{\hat M_i\}\cup \Tilde G_k\bigg)}\\
\textrm{where } &\Tilde G_k = f(\hat M, t_k)\\
& c_i \in \mathcal D\\
& \hat{m}_{VQ} = VQ(m) = c_i : i = \argmin_j{\lVert m - c_j\rVert}
\end{aligned}
\label{eq:discrete_motion}
\end{equation}
For the reconstruction task, we focused on the keyframe classification function $C(\cdot)$, vectorized quantization $VQ(\cdot)$, and transition group inpainting model $f(\cdot)$.

\subsubsection{\label{sect:keyframe}Keyframe Discovery}

For keyframe-based reconstruction, we base our initial assumption on well-established theories of human facial expression \cite{cuceloglu1972facial}. According to these theories, facial expressions consist of segments that follow a finite set of patterns. These segments can undergo transformations such as time warping, clipping, and cross-fading, which combine to produce diverse facial motion sequences.

\textbf{Assumption 1.} {A nonverbal facial motion of T frame long $\bm m$ as defined in \eqref{eq:discrete_motion}, is approximated by $k$ motion units corresponding to a sequence of latent vectors $z: z = (z_1,..., z_k)$ ($k < T$) and the $(T-k)$ transition steps between them.}

To validate this assumption, we developed a task for estimating a $k$-hot vector mask representing keyframe placement in a 3DMM facial sequence with $T$ frames. Although keyframe classification using a scoring threshold is a potential approach, the dynamic allocation of keyframes requires careful control to prevent over- or under-assignment. {More detailed discussion about alternative keyframe selection strategies can be found later in \ref{sect:fixed_sparse}}. To bypass this issue, we adopted a soft top-$k$ approach, which offers two main advantages: (1) it maintains a fixed keyframe count, resulting in a linear computational cost and mitigating crowd control concerns, and (2) it provides a flexible representation by placing keyframes densely in regions of high motion and more uniformly in stable areas.

We assumed no keyframe ground truth, which rendered the task unsupervised. First, we encoded the initial 3DMM code sequence $\bm m\in\mathbb{R}^{T\times d}$ into a latent $\bm z=(z_1,...,z_T)$ with a 1-dimensional convolution followed by a linear projection, a positional encoding, and Transformer encoder blocks. This network, denoted by $\Phi_{score}$, outputs a {channel-level} context-aware keyframe log-probability $s$, where $s_i = \Phi_{score}(\bm z, i)$, and $i$ is the frame index $i\in (1\dots T)$. After being fully trained, $s$ represents a distribution of the optimal keyframe placement from which to sample, where $s_i$ indicates the log-probability of the $i$-th frame being a keyframe; let $I$ be a discrete random variable from a $\textrm{Categorical}(p_1,...,p_n)$ distribution if $P(I=i)=p_i\quad\forall i\in N$. Then, the log-probability $s_i, i\in N$ is $\exp{s_i}\propto p_i=\frac{\exp{s_i}}{\sum_{j\in N}{\exp{s_j}}}$

\begin{equation}
    \label{eq:sample}
    I\sim \textrm{Categorical}\bigg(\frac{\exp{s_i}}{\sum_{j\in N}{\exp{s_j}}}, i\in N\bigg)
\end{equation}

By drawing $k$ largest ($\topk$) log-probability samples from \ref{eq:sample} without replacement, we obtain a subset $I = \{i_1, \dots, i_K\}$, which denotes keyframes located at frames $\{i_1,\dots,i_K\}$. Our goal is to maximize the likelihood function $\Phi_{score}$ that produces the optimal keyframe placement $I^* = \{i^*_1, \dots, i^*_K\}$, where $I^* = \arg\topk(\bm{s_i})$ with $K \leq T$, denoting the optimal keyframe placement as mentioned in \eqref{eq:i_si} to minimize the expectation of reconstruction error $\mathcal{L}_{recon}=\lVert \hat{\bm M} - \bm m\rVert$ {where $\hat{\bm M}$ is $\bm m$'s reconstruction} \eqref{eq:m_hat}. 

\begin{equation}
\label{eq:i_si}
    \begin{aligned}
s_{i}^* = \argmax_{s_{i}}{P(I=I^*)}\\
P(I=I^*)=\prod_{i=1}^{K}{\frac{\exp{s_i}}{\sum_{j=1}^{i}{\exp s_{T-j}}}}
\end{aligned}
\end{equation}

To integrate this keyframe placement learning into the gradient estimation, we utilize the Gumbel-Max trick \cite{gumbel1954statistical} along with a $\softmax$-based relaxation strategy \cite{maddison2014sampling, maddison2016concrete}. Specifically, we employ a pathwise estimator that reparameterizes the discrete random variable $I$ by separating random elements $\epsilon_i$ from deterministic components $s_i$. Specifically, we approximate $I$ using the continuous and differentiable Gumbel-Softmax approximation of the Gumbel-Max trick $i = \softmax\topk(\mathcal{G}_{s_i})$, where $\mathcal{G}_{s_i} = s_i + g_i$. Here, $g_i$ represents the i.i.d. samples drawn from the $\textrm{Gumbel}(0,1)$ distribution \cite{gumbel1954statistical}. The $\argmax$ operation is replaced by $\softmax$ to ensure differentiability \cite{maddison2016concrete}. The temperature parameter $\tau$ controls the sampling process; at low temperatures, the expected value of the Gumbel-Softmax approach is that of the categorical random variable, whereas at high temperatures, it converges to a uniform distribution over the categories \cite{jang2016categorical}:
\begin{equation}
    \label{eq:argmax}
    \arg\max_{i\in I^*}\mathcal G_{s_i}\sim \textrm{Categorical}\bigg(\frac{\exp{\frac{s_i}{\tau}}}{\sum_{j\in I^*}{\exp{\frac{s_j}{\tau}}}}, i\in I^*\bigg)
\end{equation}

{The $\topk$ selection denoted as (\ref{eq:argmax}) can be relaxed into a $k$-step iterative procedure in Algorithm \ref{alg:kf_mask} inspired by the weighted reservoir sampling \cite{vieira2014gumbel,efraimidis2006weighted}.}

\begin{algorithm}
\caption{\label{alg:kf_mask}k-hot vector for keyframe estimation}
\begin{algorithmic}[1]
\REQUIRE Logits $s$, count $K$, temperature $\tau$, length $T$.
\STATE \textcolor{gray}{\# Initialize Gumbel random variable}
\STATE $\mathbf A[\ ] \leftarrow$ empty list
\FOR{$i$ from $1$ to T}
    \STATE $g \sim Gumbel(0, 1)$
    \STATE $\mathbf{A}$.append(s[i] + g)
\ENDFOR
\STATE \textcolor{gray}{\# Iterative one-hot mask extraction}
\FOR{all $k$ from $1$ to $K$}
    \STATE $m \leftarrow \max{(1.0- \textrm{hot1}, \epsilon)}$
    \STATE $A[I] \leftarrow A[I] + \log{m}$
    \STATE \textrm{hot1} $\leftarrow \softmax(A[I] / \tau)$
    \STATE \textrm{hotK} $\leftarrow \textrm{hotK} + \textrm{hot1}$
\ENDFOR
\STATE $\textrm{hardK} \leftarrow 0$
\STATE $\textrm{idx} \leftarrow \topk(\textrm{hotk}, K)$
\STATE $\textrm{hardK} \leftarrow \textrm{scatter}$ 1 at $idx$ positions
\STATE \textcolor{gray}{\# Straight-through gradient}
\STATE $\textrm{r} \leftarrow \textrm{hardK} - sg(\textrm{hotK}) + \textrm{hotK}$
\end{algorithmic}
\end{algorithm}

In other words, given the keyframe score $s_i$ we can compute a continuous relaxation k-hot binary mask $\alpha =\sum_{i}^{K}{\alpha_i}$ where $\alpha_i^{j+1} = \alpha_i^{j}+\log{(1-p_j)}\quad \alpha^0_i\coloneqq\softmax(\mathcal G^0_{s_i})$ as illustrated in Figure \ref{fig:masking}. Finally, to optimize $s$, we determine the keyframe placement with the lowest reconstruction error $I^*$ and let $s$ approach $s^*$, where $I^*=\argmax \mathcal G_{s^*_i}$:
\begin{equation}
I^*=\argmin_{I}{\lVert\mathcal M(I,\bm m) - \bm m\lVert_2}
\end{equation}
At the start of training, $\tau$ is set to a high value to explore various keyframe placements, then gradually reduced to stabilize the optimal $\topk$ placement as the reconstruction loss converges (temperature annealing).

\subsubsection{\label{sect:fixed_sparse} Fixed $k$ versus Adaptive $k$}

{One might question why not adopt sample-wise dynamic $k$ instead of fixing it as a hyperparameter for the keyframe selection. Adaptive $k$ means the model selects the optimal $k$ that depends on the sequence's complexity. Two approaches we considered included: (i) a threshold-based method, and (ii) a sparsity-inducing penalty. Both approaches relied on interpreting the model’s output logits as binomial log-probabilities for frame selection.}

{For the threshold-based method, we experimented with differentiable smooth threshold functions (e.g., sharp sigmoid and softmax). However, this method proved unstable: as the frame scores hovered near the decision boundary, the selection oscillated significantly, leading to early stagnation in training and ultimately sub-optimal reconstructions. This, in turn, impaired the listening head prediction task.}

{In the second approach, we introduce an additional \textbf{sparsity loss} $\mathcal L_{sparse}$ to \eqref{eq:loss_function} and execute the Gumbel $\topk$ sliding kernel across temporal dimension:}
\begin{equation}
\label{eq:scarsity_loss}
    \begin{aligned}
        \mathcal L_{sparse} &= \big\lvert \sum_{c=1}^{N} p_c - K\big\rvert \\
        &\text{where } p_c =
    \begin{cases}
        1, & \text{if } c \text{ is a keyframe} \\
        0, & \text{otherwise}
    \end{cases}
    \end{aligned}
\end{equation}
{Although \eqref{eq:scarsity_loss} can encourage sparsity, joint training with the inpainting Transformer favors stable (and often degenerate) selection patterns over exploration. In most cases, we observed mode collapse, where only a single keyframe was persistently selected for most data. Although we attempted to mitigate this by carefully tuning the hyperparameter K and its weight, the problem remained prevalent across a significant portion of the training data, leading to degraded reconstruction and prediction performance. This challenge aligns with findings mentioned in \cite{hazimeh2021dselect, louizos2017learning} where an adaptive $k$ sparse representation requires a non-trivial solution. Given these difficulties, we opted to retain the fixed-k Gumbel-Top-k selection approach which demonstrated reliable and acceptable performance in both reconstruction and prediction tasks as detailed in Section \ref{sect:experiment}.} 

\subsubsection{Key Facial Motion Vectorized Quantization\label{sec:vq}}

Recent discretization-based methods such as \cite{ng2022learning,dam2024finite} utilize vector quantization encodes continuous frame-wise motion $z\in\mathbb{R}^d$ into the nearest embedding $z_c$ from a finite set of shared codebook vectors $\mathcal D\in\mathbb{R}^{d\times \lvert \mathcal D\rvert}$.  Unlike previous methods that encode dense group \cite{ng2022learning} or frame-level\cite{dam2024finite} representations, our approach leverages the learned keyframe placement in \ref{sect:keyframe}, thereby concentrating the discretization process solely on keyframe tokens. We experiment with two vector quantization implementations: VQ-VAE\cite{van2017neural} and FSQ\cite{mentzer2023finite}. According to the result found in Table \ref{tab:recon_result}, FSQ implementation achieves slightly better accuracy. While both approaches target the same codebook size, VQ-VAE requires more trainable parameters. In contrast, FSQ uses fewer parameters and tunable hyperparameters (e.g., channel number $d$ and levels $\mathcal{L}$) with a fixed-grid partitioning scheme.

\subsubsection{\label{sect:sparse2dense}Sparse-to-Dense Motion Inpainter}

Given a binary keyframe mask from Section \ref{sect:keyframe} and the quantized correspondences from Section \ref{sec:vq} of facial motion, the inpainter encodes the transition frames to reconstruct the original continuous facial motion sequence. 

The transformer-based approach has shown considerable promise, particularly in human body motion interpolation tasks like locomotion and dancing \cite{mo2023continuous, duan2021single}. However, these methods often assume periodic latent structures and rigid skeletal constraints, with clearly defined keyframes. In contrast, our study deals with the more complex features of facial expressions, which emerge from 3D face shape reconstruction tasks. Here, facial muscle activation, along with changes in expression and pose, is learned by optimizing the deformation of a template face model to match a 2D appearance \cite{tellamekala20233d, feng2021learning, wang2022faceverse}. To tackle this complexity, we utilized an established keyframe-based context from previous sections. Our inpainter design is a transformer network, denoted as $\phi_{full}$, which performs inter- and extrapolation to predict the intermediate frames between keyframes. This is achieved based on the vectorized, quantized representation $z^{kf} \in \mathbb{R}^{T \times D}$ and the set of keyframe indices $T^{kf} = {i_1, \dots, i_K}$, along with their corresponding relative positions.
\begin{equation}
\label{eq:m_hat}
\hat{\bm M} = \phi(z^{kf}, T^{kf})
\end{equation}

\textbf{Assumption 2.} Given that we blank out 3DMM features from the transition frames, we hypothesize that we can recover these in-between states $\mathbf m^{tf}$ with the given $k$ keyframes $\mathbf m^{kf}$ and their respective indices $\mathbf T^{kf}\in\mathbb{R}^k$ and $T^{tf}\in\mathbb R^{N-k}$.

To verify this assumption, we prepared a transformer architecture \cite{vaswani2017attention} whose main components include two transformer encoders (Keyframe encoder and transition frame encoder) and one decoder. The first encoder, the keyframe encoder, encodes 3DMM features at keyframes into latent $z^{kf}$. The transition frame encoder $\phi^{tf}$ converts $(T^{tf},z^{kf})$ into the transition frame latent vectors $z^{tf}$. Finally, the decoder combines all the intermediate variables $z^{kf}$, $T^{kf}$, $z^{tf}$, $T^{tf}$ to generate the reconstruction {$\hat{\bm M}$}. 
\begin{equation}
    \hat{\bm M} = \Phi^{full}(\mathbf z^{kf}, \mathbf T^{kf}, \Phi^{tf}(\mathbf z^{tf}_{pe}, \mathbf z^{kf}), \mathbf T^{tf}) 
\end{equation}
Both encoders are built on multiple encoder layers, each with a multihead self-attention layer and a feedforward network. Specifically, a transitional encoder utilizes sinusoidal positional encoding (PE) \cite{vaswani2017attention, mo2023continuous} to transform $T^{tf}$ into a binary sliding vector $z^{tf}_{PE} = PE(T^{tf})$ for concatenation with $z^{kf}$ to form a positional embedding. This concatenated PE addresses the sensitivity to minor changes in the embedding of 3DMM features \cite{tellamekala20233d} that occur with additive PE. The fused information is then leveraged as query $Q$ for the attention layers to output $z^{tf}$, which reflects the temporal difference from the surrounding keyframe $z^{kf}$.

\subsubsection{ Discussion of Sparse Embedding Information Loss}
The sparse approach offers two advantages: first, it disentangles locally dependent facial motion patterns before learning, thereby enhancing the clarity of encoded quantized expression signals \cite{le2016sparsity}; second, it acknowledges that nonverbal communication encompasses both controlled and involuntary facial expressions \cite{buck1994social}, thus enabling the construction of facial motion in a more flexible and human-friendly manner \cite{buck1994social}. A primary concern with sparse representation is whether the excluded information can adequately capture the richness of facial motion related to expressions. To quantitatively assess the robustness and contribution of the proposed sparse representation approach to improving reconstruction accuracy, we conducted a comparison between the dense and sparse representation methods. Additionally, to evaluate the effectiveness of the dynamic keyframe approach, we performed experiments across various settings, including quantization methods, keyframe number, and keyframe strategies, as shown in Tables \ref{tab:dynamic_result} and \ref{tab:recon_result}.

\subsection{Sparse Multimodal Listening Head Prediction}
In the listening head prediction task, we incorporated the learned sparse facial motion structure into a predictive model. This model, which utilizes the multimodal conversational context from both the listener and speaker, generates contextually appropriate feedback based on a trained dataset. In formal terms, given a dyadic conversation with video and audio components, we generate listener feedback at frame $k^{th}$, denoted as $\mathbf m^{L}_k$, conditioned on the listener's previous feedback $\mathbf m^L_{<k}$ and the speaker's facial expressions $\mathbf m^S$ and audio $\mathbf a^S$. By leveraging the finite quantized codebook $D$ learned in Section \ref{sec:sparse}, which consists of $M$ observed facial tokens $z^D_1, \dots,z^D_M$, we model the probability distribution of the predicted expression $\mathbf{m}^L_k$ at the $k^{th}$ frame in an autoregressive manner. This effectively transforms the problem into a next-token prediction task, which is a well-established sequence modeling task:
\begin{equation}
\begin{aligned}
    \textrm{Pr}(\mathbf m^L_k, ..., \mathbf m^L_{0}) = \textrm{Pr}(\mathbf m^L_k)\prod^{k-1}_{n=0}\textrm{Pr}(\mathbf m^L_n\lvert \mathbf m^L_{<n}, \mathbf m^S_{\leq n}, \mathbf a^S_{\leq n})
\end{aligned}
\end{equation}
In contrast to standard facial motion token generation, our approach predicts two distinct types of tokens for each future frame: a keyframe token and a transition token. Once the model is fully trained, the sparse structure introduced in \ref{sect:keyframe} employs a $\topk$ strategy, rendering the keyframe placement estimation process deterministic. This allows the prediction task to be self-supervised by utilizing the keyframe mask and discrete token codebook generated by the reconstruction task described in Section \ref{sec:sparse}. These derived labels serve as the target and past context for the listeners -facial motion prediction. 

In contrast to recent studies such as \cite{liu2024one, song2023react2023}, which discarded past listener facial expressions owing to noise introduced by distribution shifts between the training phase (with ground truth) and the inference phase, we opted to retain this modality. This decision was made to model the listener's intention more accurately, as supported by previous research \cite{poppe2013perceptual, ng2022learning, song2023multiple}. To address the distribution shift problem that arises at the start of the inference when no ground truth is available, we employed an augmentation technique that prepends a neutral sequence to the beginning of the training sequence.

\begin{figure}[t!]
\centering
\includegraphics[width=0.5\textwidth, trim={5cm 6cm 4.5cm 0cm}, clip]{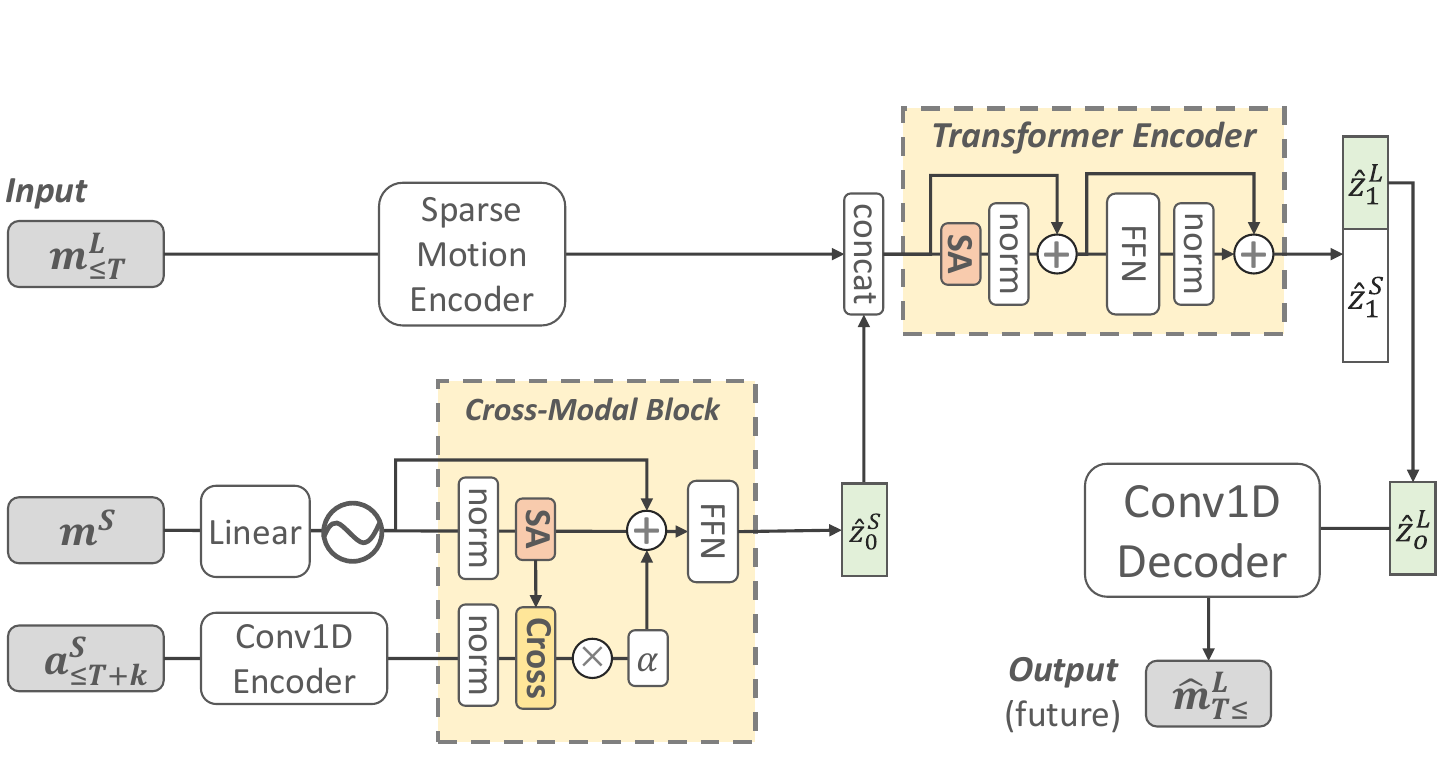}
\caption{\label{fig:predictor_arch} \textbf{Predictor's architecture overview.} A transformer-based predictor is employed for the next-token prediction task based on multi-modal input context in a dyadic conversation.}
\end{figure}
\subsubsection{Speaker's Multimodality Context Fusion}
In addition to the listener's visual context, our predictor leverages the speaker's visual and audio contexts for prediction. Speech tokens encoded by Wav2Vec 2.0 \cite{baevski2020wav2vec} were fed directly into a multimodal speaker encoder without modification. However, for continuous high-dimensional audio features, as in the L2L scenario discussed in Section \ref{sec:prelim}, a Conv1D feature extractor followed by a max-pooling layer is employed to align the temporal dimensions of the visual and audio modalities. 

Cross-modal fusion is not a new problem; however, the debate regarding its solution remains divided and open, with studies such as \cite{zhan2023multimodal, xu2023multimodal} concluding that the best solutions are task- and data-specific.  In our investigation, we implemented a cross-attention mechanism inspired by \cite{ng2022learning}, but with a more recent gating dual encoder, as proposed by \cite{dou2022coarse}. This approach efficiently fuses a speaker's facial expression $x$ and speech $y$ into an intermediate latent embedding $\mathbf{z}^S_0$ (see Figure \ref{fig:predictor_arch}). The fusion process is governed by a learnable parameter $\alpha$, which dynamically modulates the relative contributions of the two input streams to the fused embedding.
\begin{equation}
    \begin{aligned}
\Tilde{x} &= \textrm{Self-Att}(x)\\
x &= x + \Tilde{x} + \alpha\times \textrm{Cross-Att}(\Tilde{x},y)\\
x &= x + \textrm{FFN}(x)
\end{aligned}
\end{equation}
\subsubsection{Speaker-Listener Context Encoder/Decoder}

As shown in Figure \ref{fig:predictor_arch}, the architecture consists of two encoders---one for the listener and one for the speaker---and one for the decoder. Past listener and speaker context latents are concatenated and fed as a query into the transformer encoder stack \cite{vaswani2017attention}. The resulting intermediate embedding is decoded into the prediction's 3DMM parameters. The decoding process is modeled as a multiclass prediction aligned with the token-like nature of the listener's sparse facial motion representation, where transition frames are marked as the 0 class.
 
\begin{table*}
\centering
\caption{\label{tab:pred_result1}Comparison of our approach with Learning2Listen\cite{ng2022learning} baseline test set.}
    \begin{tabular}{llllllllllllll} 
\toprule
 & \multicolumn{6}{c}{\textbf{Expression }} &  & \multicolumn{6}{c}{\textbf{Rotation }} \\ 
\cmidrule{2-7}\cmidrule{9-14}
 & \multicolumn{2}{l}{\textbf{Appropriateness}} & \multicolumn{2}{l}{\textbf{Diversity}} & \multicolumn{2}{l}{\textbf{Synchrony}} &  & \multicolumn{2}{l}{\textbf{\textbf{\textbf{\textbf{Appropriateness}}}}} & \multicolumn{2}{l}{\textbf{\textbf{Diversity}}} & \multicolumn{2}{l}{\textbf{\textbf{\textbf{\textbf{Synchrony}}}}} \\ 
\cmidrule{2-7}\cmidrule{9-14}
  & \textbf{L2~}($\downarrow$) & \textbf{FD~}($\downarrow$) & \textbf{Var}($\cdot$) & \textbf{SI}($\cdot$) & \textbf{P-FD}($\downarrow$) & \textbf{RPCC}($\downarrow$) & &\textbf{\textbf{L2~}}($\downarrow$) & \textbf{\textbf{FD~}}($\downarrow$) & \textbf{\textbf{Var}}($\cdot$) & \textbf{\textbf{SI}}($\cdot$) & \textbf{\textbf{P-FD}}($\downarrow$) & \textbf{\textbf{RPCC}}($\downarrow$) \\ 
  &&\scriptsize{[$\times 10^3$]} &&&\scriptsize{[$\times 10^3$]}&\scriptsize{[$\times 10^{-1}$]}&&&\scriptsize{[$\times 10^2$]} &&&\scriptsize{[$\times 10^2$]}&\\
\midrule
Ground truth & - & 0.00 & 2.90 & 2.61 & - & - &  & - & - & 0.81 & 1.96 & - & -\\ 
\midrule
Random & 129.34 & 524.69 & 62.23 & 1.17 & 526.46 & 0.8 &  & 27.67 & 257.06 & 62.39 & 1.06 & 257.16 & 0.002 \\
Median & 43.18 & 97.86 & 0.0000 & 0.000 & - & - &  &  &  &  &  &  &  \\ 
\midrule
LFI \cite{jonell2020let} & 50.07 & 43.63 & 1.15 & 1.33 & 54.34 & 8.0 &  & 9.00 & 9.80 & 0.17 & 1.07 & 12.36 & 0.034 \\
Learning2Listen \cite{ng2022learning} & 33.16 & 3.55 & 2.01 & 2.48 & 5.15 & 0.2 &  & 4.75 & 0.81 & 0.62 & 1.82 & 0.87 & \textbf{0} \\
ELP \cite{song2023emotional} & - & 1.37 & \textbf{2.70} & 2.15 & - & 0.14 & &-&\textbf{0.36} & 0.59 & 1.60 & - & 0.077 \\
\midrule
{\cellcolor[rgb]{0.855,0.91,0.988}}\textbf{Ours} & {\cellcolor[rgb]{0.855,0.91,0.988}}\textbf{26.65} & {\cellcolor[rgb]{0.855,0.91,0.988}}\textbf{1.13} & {\cellcolor[rgb]{0.855,0.91,0.988}}2.17 & {\cellcolor[rgb]{0.855,0.91,0.988}}\textbf{2.63} & {\cellcolor[rgb]{0.855,0.91,0.988}}\textbf{1.35} & {\cellcolor[rgb]{0.855,0.91,0.988}}\textbf{0.023} & {\cellcolor[rgb]{0.855,0.91,0.988}} & {\cellcolor[rgb]{0.855,0.91,0.988}}\textbf{4.02} & {\cellcolor[rgb]{0.855,0.91,0.988}}0.68 & {\cellcolor[rgb]{0.855,0.91,0.988}}\textbf{0.83} & {\cellcolor[rgb]{0.855,0.91,0.988}}\textbf{2.03} & {\cellcolor[rgb]{0.855,0.91,0.988}}\textbf{0.73}  & {\cellcolor[rgb]{0.855,0.91,0.988}}0.006 \\ 
\bottomrule
 &  &  &  &  &  &  &  &  &  &  &  &  &  \\
\multicolumn{14}{l}{($\cdot$) means the closer to the ground truth, the better.} \\
\multicolumn{14}{l}{\:-\: denotes the left out measurements from the office report.}\\
\multicolumn{14}{l}{\textbf{Bold} metric indicates the best performance for a metric.}\\
{\cellcolor[rgb]{0.855,0.91,0.988}}\textbf{Colored bold row}&\multicolumn{13}{l}{indicates the technique with the best overall performance.}\\
\end{tabular}
\end{table*}

\begin{table*}[t]
\centering
\caption{\label{tab:pred_result2}Comparison of our approach with one-to-many baselines on REACT \cite{song2023react2023} test set.}
\begin{tabular}{llllllll} 
\toprule
 & \multicolumn{2}{l}{\textbf{Appropriateness}} & \multicolumn{3}{l}{\textbf{Diversity}} & \textbf{Realism} & \textbf{Synchrony} \\ 
\cmidrule(l){2-8}
 & \textbf{FRCorr} ($\uparrow$) & \textbf{FRDist} ($\downarrow$) & \textbf{FRDiv} ($\uparrow$) & \textbf{FRVar} ($\uparrow$) & \textbf{FRDvs} ($\uparrow$) & \textbf{FRRea} ($\downarrow$) & \textbf{FRSyn} ($\cdot$) \\ 
\midrule
Ground truth & 0.85 & 0.00 & 0.0000 & 0.0724 & 0.2483 & 82.45& 47.69 \\ 
\midrule
Random & 0.05 & 237.23 & 0.1667 & 0.0833 & 0.1667 & - & 44.10 \\
Mime & 0.38 & 92.94 & 0.0000 & 0.0724 & 0.2483 & - & 38.54 \\
MeanFr & 0.00 & 97.86 & 0.0000 & 0.0000 & 0.0000 & - & 49.00 \\ 
\midrule
Trans-VAE & 0.07 & 90.31 & 0.0064 & 0.0012 & 0.0009 & 69.19 & 44.65 \\
BeLFusion & 0.12 & 94.09 & 0.0379 & 0.0248 & 0.0397 & 94.09 & 49.00 \\
Dense-FSQ \cite{dam2024finite} & 0.31 & 84.93 & 0.1164 & 0.0348 & 0.1166 & \textbf{34.66} & 47.42 \\
\hline
\rowcolor[rgb]{0.855,0.91,0.988} \textbf{Ours} & \textbf{0.84} & \textbf{66.89} & \textbf{0.1207} & \textbf{0.0871} & \textbf{0.1212} & 35.78 & \textbf{45.66} \\ 
\bottomrule
 &  &  &  &  &  &  &  \\
\multicolumn{8}{l}{($\cdot$) means the closer to the ground truth, the better.} \\
\multicolumn{8}{l}{indicates the best average performance among the heuristic baselines for the groups of metrics.}
\end{tabular}
\label{tab:resulttable}
\end{table*}
 
\subsection{Training}
\subsubsection{Loss Function for Sparse Representation Learning}

We simultaneously trained the joint keyframe logits embedding, keyframe vector quantization, and transition frame inpainting tasks using two loss components: motion loss $\mathcal{L}_2$, quantization loss $\mathcal{L}_2^{kf}$, and masking loss $\mathcal{L}_1^{mask}$.
\begin{equation}
    \begin{aligned}
    \mathcal L &= \mathcal L_2 + \mathcal L_2^{Q} + \alpha \mathcal L_1^{mask} \\
    &= \lVert m - \hat m\rVert + \lVert z^{kf} - \hat z^{kf} \rVert \\
    &+ \lVert \textrm{sg}[E(z^{kf})] - z^{kf}_q\rVert + \lVert \textrm{sg}[z^{kf}_q] - E(z^{kf})\rVert_2\\
    &+ \alpha \sum_c^C{\lvert p_c - p_{max}\rvert}
\end{aligned}
\label{eq:loss_function}
\end{equation}

Motion loss focuses on the full sequence, whereas quantization loss \cite{van2017neural} mitigates reconstruction errors during keyframe quantization. Optional masking loss is applied during $\topk$ sampling to adjust the framewise attention scores and optimize them for the most accurate sequence reconstruction.

Reasoning-wise, motion loss is straightforward for the task \cite{ng2022learning, song2023react2023, dam2024finite}; however, the other two terms are novel for the facial motion reconstruction task to cope with the sparse structure and overall training design. It is noteworthy that although the straight-through trick allows the gradient to pass through the non-differentiable, the quantization requires a reconstruction loss itself at the beginning and can be focally adjusted to only the keyframes marked by the $\topk$ operator later on.

\subsubsection{Joint Sparse Embedding Training Strategy}
In our experience, naively training the joint task directly led to unstable loss convergence because the keyframe logits tended to move toward suboptimal solutions, whereas the quantization embeddings were still unstable. To address this issue, we employ two optimizers to train the keyframe log-probability scores and the rest of the network respectively. Without the additional optimizer, the loss converges shortly and fluctuates drastically showing no sign of improvement.

\begin{figure}[t]
\centering
\includegraphics[width=0.5\textwidth, trim={6.5cm 3cm 6.5cm 3.5cm}, clip]{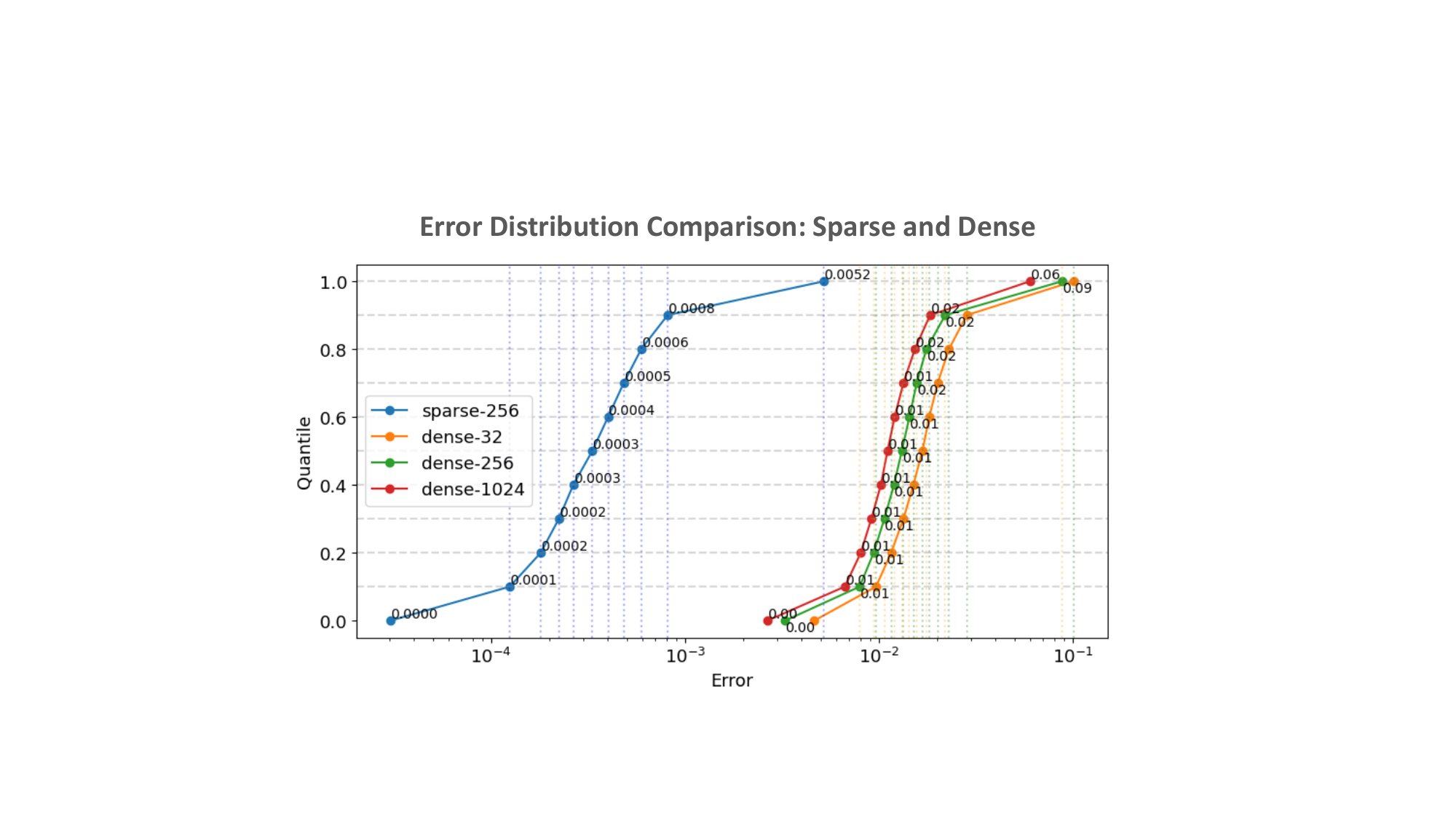}
\caption{\label{fig:sparse_dense_ablation} {\textbf{Quantile distribution of log-scaled error between sparse and dense representations.} The x-axis shows the log-transformed absolute error, while the y-axis represents quantile levels.}}
\end{figure}

\subsubsection{Listening Head Future Token Prediction}

As mentioned earlier, we modeled the listening head pose and motion prediction as a next-token prediction problem. The objective loss comprises the cross-entropy and binary soft dynamic time-warping functions. During training, we employ a teacher-forcing scheme using ground-truth-encoded tokens. In our experiments with sparse representation, training with a single future token resulted in poor keyframe token recall. This issue is expected given the sparse structure, where most tokens are non-keyframes, unlike natural language processing (NLP) tasks where the token distribution is typically balanced. We take into account this problem with balancing weights $w$ to the loss formula. The cross-entropy $CE$ maximizes the probability of the ground truth token $t\in \mathcal V\coloneqq\{1,...,V\}$ dictionary within every sequence of our input:

\begin{equation}
\textrm{CE}(\hat t, t) \coloneqq \frac{1}{\mathcal T}\sum_{i=0}^{\mathcal T}{-w_i\log{(\mathcal P(t_{i}\lvert t_{< i})}}
\end{equation}

\section{Experimental Result}
\label{sect:experiment}

In this section, we describe the experimental setup (Sections \ref{sec:recon_setup}\&\ref{sec:pred_setup}). To demonstrate the effectiveness of the sparse structure, we compared the performance of the proposed methods with state-of-the-art solutions in the reconstruction and listening head token prediction task.

\subsection{Implementation Details}

We used the AdamW \cite{loshchilov2017decoupled} optimizer with $\beta_1 = 0.9$ and $\beta_2 = 0.98$, with the cosine annealing\cite{loshchilov2016sgdr} as the learning rate scheduler to train both tasks on a 1x NVIDIA GeForce 4060 graphics processing unit (GPU) for 2 to 4 hours on average.

\begin{figure*}[t!]
\centering
\includegraphics[width=1.0\textwidth, trim={0cm 6cm 0cm 0.5cm}, clip]{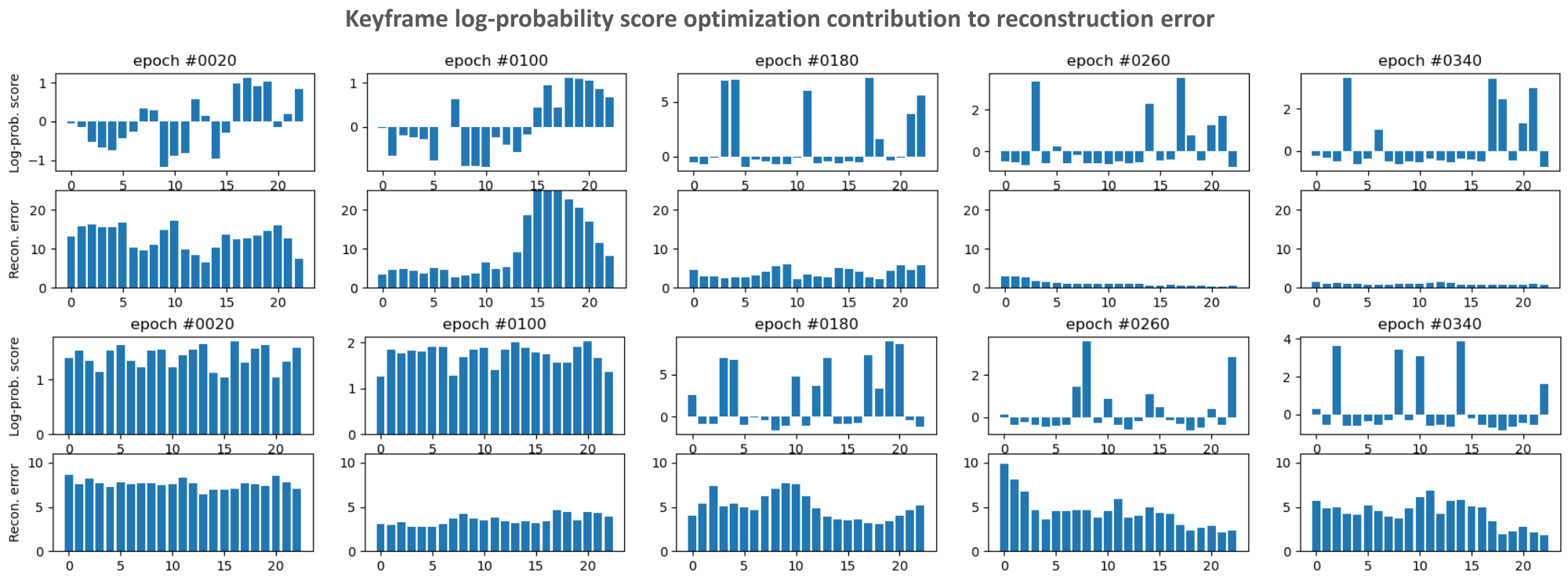}
\caption{\label{fig:keyframe_error} \textbf{Illustration of reconstruction error development during training}. Two examples extracted from keyframe log-probability score (k=7) training. \textbf{Top}---As the reconstruction error distributes mostly at the end, the log probability focuses on the last few frames. At later epochs, more keyframes are assigned for the first half sequence to balance the error. \textbf{Bottom}---As the error is distributed evenly, the keyframe assignment converges to the uniform placement.}
\end{figure*}

\subsection{\label{sec:recon_setup}Sparse Motion Reconstruction}

The reconstruction task converts continuous facial motion sequences into discrete tokens, with keyframes represented by codebook vector indices and transition frames encoded with positional and keyframe-wise information. We show that this setup effectively represents facial motion. The reconstruction window was set to 48, considering the short-duration L2L dataset (64-frame long \cite{ng2022learning}). All architecture modules used a 256-dimensional embedding, except the first and last projection layers, optimized for our GPU's video random access memory (VRAM).

To verify the contribution of the novel dynamic keyframe setting, we implemented an additional uniform keyframe and VQ-only baseline. The result of the comparative analysis of the reconstruction loss between the proposed technique and the baseline can be found in Table \ref{tab:dynamic_result}.
\begin{enumerate}
    \item \textbf{VQ-only}: {all frames are vector-quantized}\cite{dam2024finite}.
    \item {\textbf{Static}: keyframes are uniformly sampled at every $\floor{\frac{N}{k}}$}.
    \item {\textbf{Dynamic}: our proposed $\topk$ strategy.}
\end{enumerate}

\begin{table}[t]
\centering
\caption {\label{tab:dynamic_result}Ablation study on dynamic keyframe contribution}
\begin{tabular}{lllll} 
\toprule
 &  \multicolumn{4}{c}{\textbf{MSE} \scriptsize{[$\times 10^{-2}$]}} \\ 
\cmidrule(l){2-5}
 & \multicolumn{1}{c}{\textbf{k=3}} & \textbf{k=7} & \textbf{k=10} & \textbf{All}\\ 
\hline
\\
VQ-only (no keyframe) \cite{dam2024finite} & - &  - & - & 0.26\\
Static (Ours) & 0.89 & 0.64 & 0.33 & -  \\
{\cellcolor[rgb]{0.855,0.91,0.988}}\textbf{Dynamic} (Ours) & {\cellcolor[rgb]{0.855,0.91,0.988}}0.15 &  {\cellcolor[rgb]{0.855,0.91,0.988}}0.13 & {\cellcolor[rgb]{0.855,0.91,0.988}}0.12 & {\cellcolor[rgb]{0.855,0.91,0.988}}-\\
\hline
\end{tabular}
\end{table}
\raggedbottom

The results listed in the ablation demonstrate that dynamic keyframe placement significantly outperforms the two baselines in terms of reconstruction error across all $k$ hyperparameter settings. The joint training scheme enables our model to simultaneously learn both optimal keyframe placement and other reconstruction modules, effectively adapting to varying and challenging input facial motion patterns as shown in Figure \ref{fig:keyframe_error}. {To further demonstrate the effectiveness of our sparse representation in preserving high-fidelity facial behavior, we conduct an ablation study shown in Figure \ref{fig:sparse_dense_ablation}. We compare reconstruction performance between our sparse representation ($|C|=256$) and dense codebooks of varying sizes ($|C|=32$, $200$, and $1024$). The sparse approach achieves up to two orders of magnitude lower error, attributed to the combination of keyframe identification and motion inpainting. Unlike dense methods that apply tokenization across all channels, often disrupting temporal coherence and inter-channel dependency \cite{xie2024g2p}. For this reason, dense methods typically avoid per-channel quantization, opting instead for whole-timestep \cite{dam2024finite}, grouped \cite{ng2022learning}, or hierarchical \cite{xie2024g2p} strategies. On the contrary, our approach performs channel-wise tokenization with independent keyframe allocation, offering greater flexibility while preserving inter-channel dependencies. While larger dense codebooks and bigger datasets may narrow the gap in the future, challenges such as low codebook utilization and high computational cost remain as potential selling points for sparse and small-sized codebook representations.} 

\begin{figure*}[t!]
\centering
\includegraphics[width=1.0\textwidth, trim={3cm 3cm 4cm 4cm}, clip]{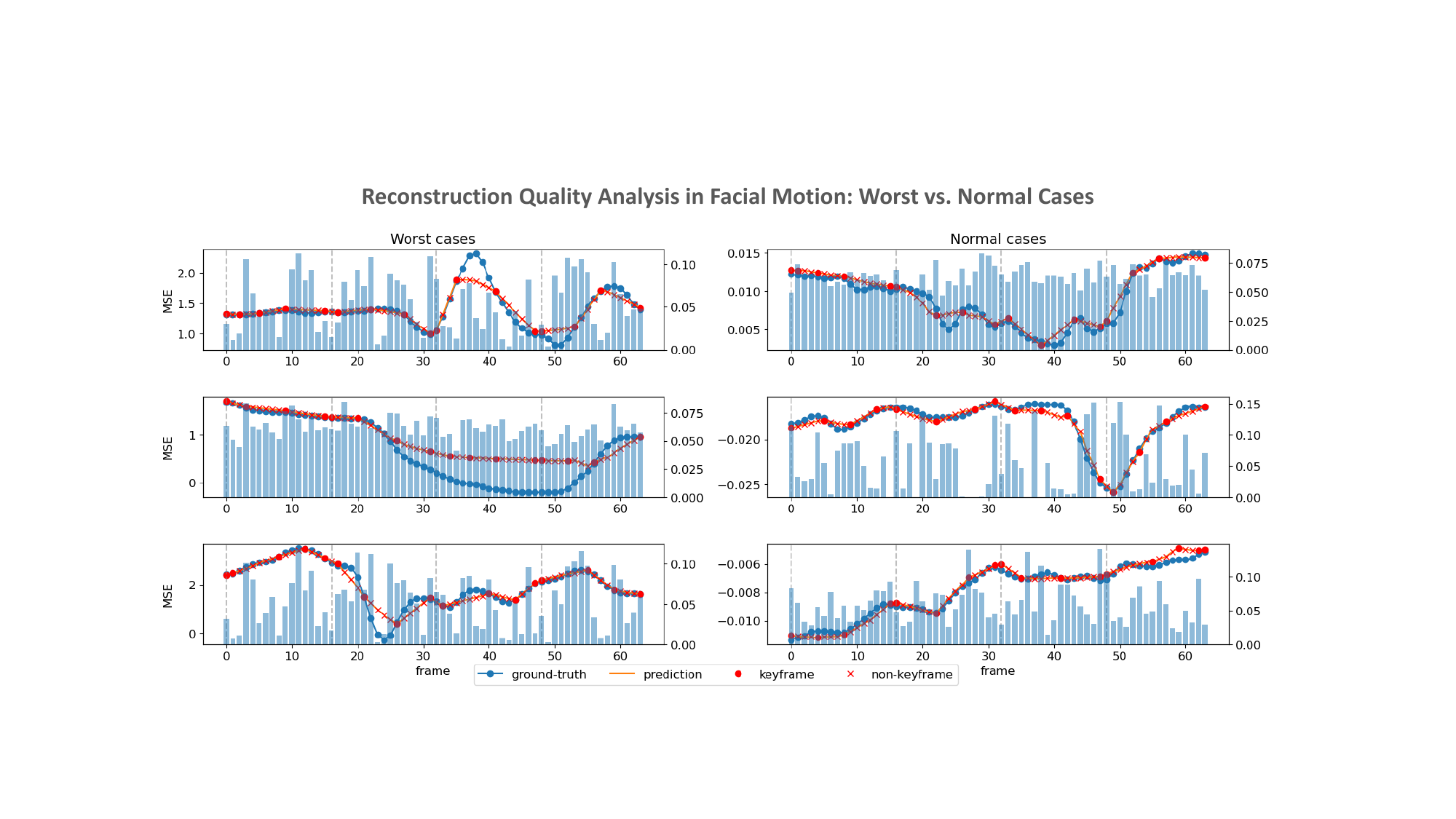}
\caption{\label{fig:detailed_error} {\textbf{Visualization of facial motion reconstruction quality across multiple samples, highlighting worst-case (left) and normal-case (right) behaviors}. Each subplot presents channel-wise mean squared error (MSE, bars) and predicted vs. ground-truth trajectories (lines), along with keyframe (red circles) and non-keyframe (red x) probabilities.}}
\end{figure*}

{Finally}, we verified the trade-off between the codebook size, quantization technique, and reconstruction error between dense and sparse structures. Our quantitative evaluation of the reconstruction task used four candidates for comparison.
\begin{enumerate}
    \item \textbf{L2L}: Group-based discrete tokenization \cite{ng2022learning}
    \item \textbf{Dense-VQ}: Our L2L revision with a VQ-VAE\cite{dam2024finite} 1024-element codebook.
    \item \textbf{Sparse-VQ}: Our sparse structure with a VQ-VAE.
    \item \textbf{Sparse-FSQ}: Our sparse structure with FSQ.
\end{enumerate}

\begin{table}[t]
\centering
\caption {\label{tab:recon_result}Comparison of facial motion reconstruction error}
\begin{tabular}{lllll} 
\toprule
 &  &  & \multicolumn{2}{c}{\textbf{MSE}} \\ 
\cmidrule(l){4-5}
 & \textbf{\textbf{Variant}} & \textbf{\#Params} & \multicolumn{1}{c}{\textbf{L2L}} & \textbf{REACT23} \\ 
 &&&\scriptsize{[$\times 10^{-2}$]}&\scriptsize{[$\times 10^{-2}$]}\\
\hline
\\
L2L \cite{ng2022learning} & VQ-256 & \multicolumn{1}{r}{13.0 M} & 1.44 &  17.91\\
Dense-FSQ \cite{dam2024finite} & FQ-1024 & \multicolumn{1}{r}{13.2 M} & 0.73 & 2.12  \\
Sparse-VQ (Ours) & VQ-256 & \multicolumn{1}{r}{9.6 M} & 0.05 &  0.80\\
{\cellcolor[rgb]{0.855,0.91,0.988}}\textbf{Sparse-FQ (Ours)} & {\cellcolor[rgb]{0.855,0.91,0.988}}FSQ-256 & \multicolumn{1}{r}{{\cellcolor[rgb]{0.855,0.91,0.988}}9.6 M} & 0.03 {\cellcolor[rgb]{0.855,0.91,0.988}} & 0.79{\cellcolor[rgb]{0.855,0.91,0.988}}\\
\hline
\end{tabular}
\end{table}

\begin{figure*}[t!]
\centering
\includegraphics[width=0.9\textwidth, trim={0cm 0cm 0cm 0cm}, clip]{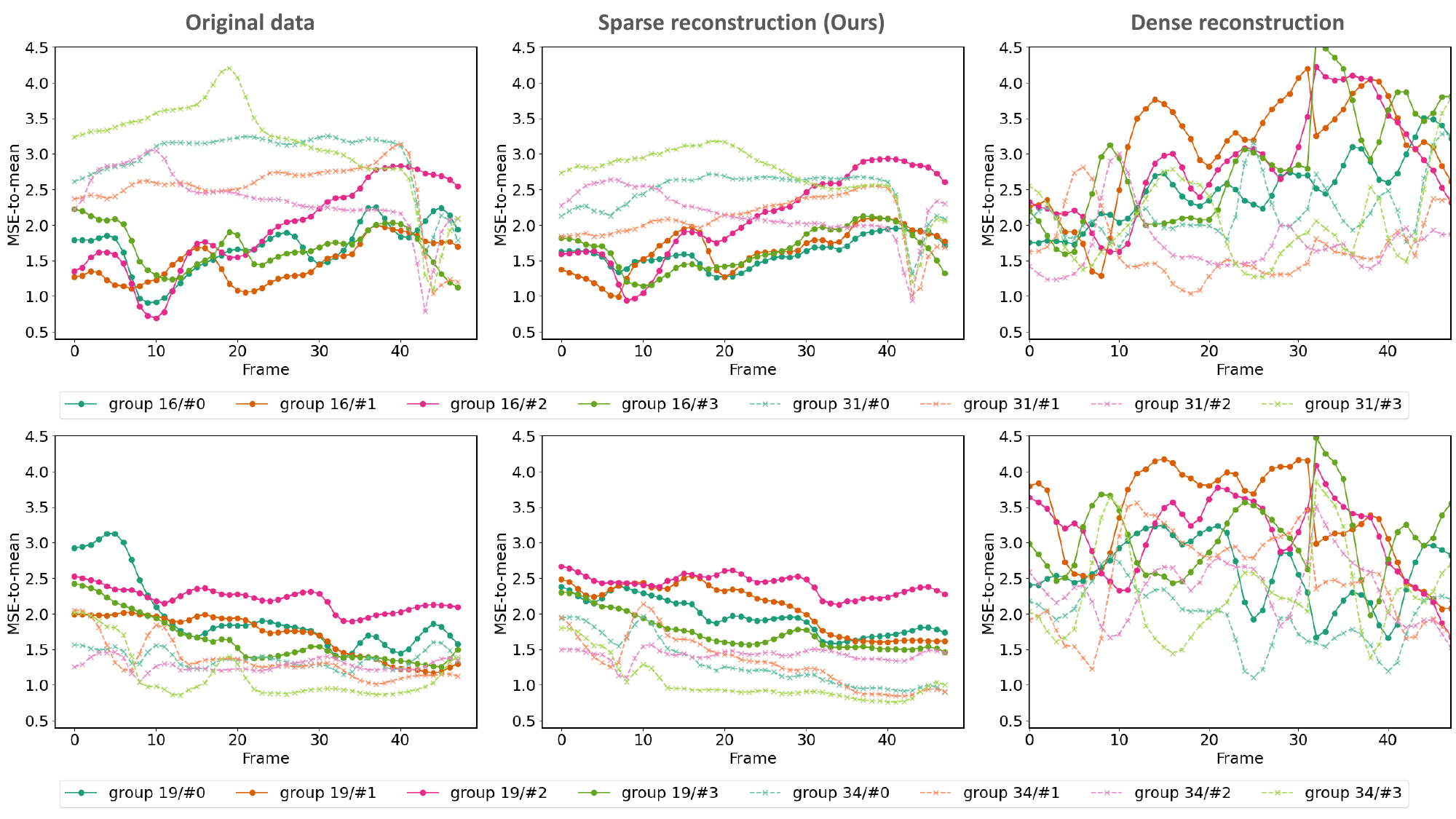}
\caption{\label{fig:noise_pattern} \textbf{Visual comparison on cluster integrity on the Learning2Listen dataset}. L2L's comparative temporal cluster integrity. We plot two groups of highly distinctive facial motions before and after encoding the temporal-wise error from the corresponding centroid. A better cluster integrity preservable technique maintains discernible distances between upper and lower clusters after the discretization process. Overall, our sparse representation preserves the temporal and cluster integrity structure better. }
\end{figure*}

The empirical results in Table \ref{tab:recon_result} show an improvement in the trade-off. The FSQ layer performs slightly better with a more efficient design, leading us to adopt the FSQ quantizer for the prediction task. {As \textbf{SFMS} is especially good at representing continuous facial motion, sparsely maintaining good balance between continuity and accuracy. Although the sparse structure is learned by the reconstruction task, the differentiable sampling process design makes the estimated keyframes log-probability clustering to high variance regions and overly neglects other regions. While a fixed number keyframe window may sound limited as a representation compared to the dynamic number of keyframes, we found the latter tends to be more unstable to train and therefore less accurate for the reconstruction task, which aligns with previous discussion in \ref{sect:fixed_sparse}. Due to the early poor facial motion reconstruction accuracy, we did not mention the result in the experiment. We include the channel-wise analysis on the reconstruction for keypoint estimation capability demonstration in Figure \ref{fig:detailed_error}.}

{Figure \ref{fig:sparse_dense_ablation} shows that despite outperforming the dense baselines with a smaller size codebook, our proposed sparse representation’s highest error quantiles (top 10\%) show a slightly steeper deterioration in reconstruction accuracy. This highlights a potential limitation of sparse representations: in rare cases, suboptimal or insufficient keyframe allocation can significantly impact reconstruction quality, as the model must interpolate transitions in high-dimensional motion spaces. Two primary failure modes were observed: (i) codebook underfitting due to vector quantization, where facial features are poorly represented (e.g., Left Case \#2), and (ii) misaligned keyframe selection (e.g., Left Cases \#1 and \#3), which can cause the model to diverge from the intended motion pattern. Additional qualitative examples are provided in Figure \ref{fig:detailed_error}. It is important to note that vector quantization failures also affect dense representations. In fact, dense models tend to be more sensitive to these issues due to the over-representation of redundant transitions in facial expression distributions. }

{Regarding keyframe selection, we observed that even minor misplacements (e.g., Left \#1 and Right \#1 in Figure 4) can lead the inpainting module to overlook high-frequency motion breakpoints. This behavior stems from the nature of the differentiable log-probability scores produced by our proposed LogitsEncoder: (i) the inherently smooth distribution tends to allocate more keyframes to regions with sharp but simple peaks, potentially neglecting more complex but subtler transitions, and (ii) in sequences with high-frequency or complex dynamics, small shifts in keyframe predictions (1–2 frames) can cause multiple local patterns to merge, oversimplifying the representation. While we experimented with temperature annealing to sharpen the log-probability distribution and reduce misassignments, this approach degraded performance on other sequence types. This suggests that a more sophisticated and adaptive keyframe selection mechanism may be required in the future to resolve this problem. }

Although these limitations do not significantly impact the reconstruction or prediction tasks, as confirmed by our experimental results, we believe that enhancing codebook expressiveness and enabling dynamic keyframe selection are promising directions for further improving the proposed sparse representation framework.

\subsection{\label{sec:pred_setup} Sparse Listening Head Prediction}
For training, we generated mini-batches of sliding past-future windows: a 40-frame-long past context, an 8-frame-long future window. During training, the predictor is trained on parallel prediction on 8-frame-long future windows; each predicting target is either a transition frame or the learned codebook.

During inference, the autoregressive method rolls the prediction window into the past context for the next token prediction to obtain the final results. The photorealistic visualization was generated using ROME\cite{khakhulin2022realistic} and PIRender\cite{ren2021pirenderer} for DECA and FaceVerse prediction, respectively.


\subsection{Comparison With State-of-the Art Methods}

We compare our \textbf{SFMS} with the state-of-the-art listening head prediction methods including Let's Face it \cite{jonell2020let}, L2L \cite{ng2022learning}, Emotional Listener Portrait (ELP) \cite{song2023emotional}, DenseFSQ\cite{dam2024finite}, Behavioral Latent difFusion (BeLFusion)\cite{song2023react2023}, and Trans-VAE \cite{song2023react2023}. We demonstrate the effectiveness of our proposed method on two fronts: first, \textbf{SFMS} sparse discrete tokens mitigate motion temporal dynamics and diversity loss during the encoding process; second, \textbf{SFMS} and the sparse predictor improve the listening head prediction objective compared to others.

\subsubsection{Quantitative Comparison}

For the first hypothesis, we tested two criteria: generalized accuracy and distinct pattern disentanglement ability between the dense and sparse motion discretization approaches. The first analysis tests the cluster integrity, which is one of the problems that we believe has limited previous work, where distinct pattern group signals are mixed together because they have to overfit both key and transition facial motions under the same codebook. We first converted the test facial motions into lossy reconstructed motions and then compared the cluster integrity both quantitatively (via cluster evaluation metrics) and qualitatively (via visual inspection; see Figure \ref{fig:noise_pattern}). \textbf{SFMS} lightens the burden over the quantization codebook by only learning key motion states and letting the inpainting network interpolate or extrapolate the transition phase, thereby mitigating the motion structure loss without increasing the codebook size.

\begin{figure*}[ht!]
\centering
\includegraphics[width=0.8\textwidth, trim={1.5cm 0.2cm 1.5cm 0cm}, clip]{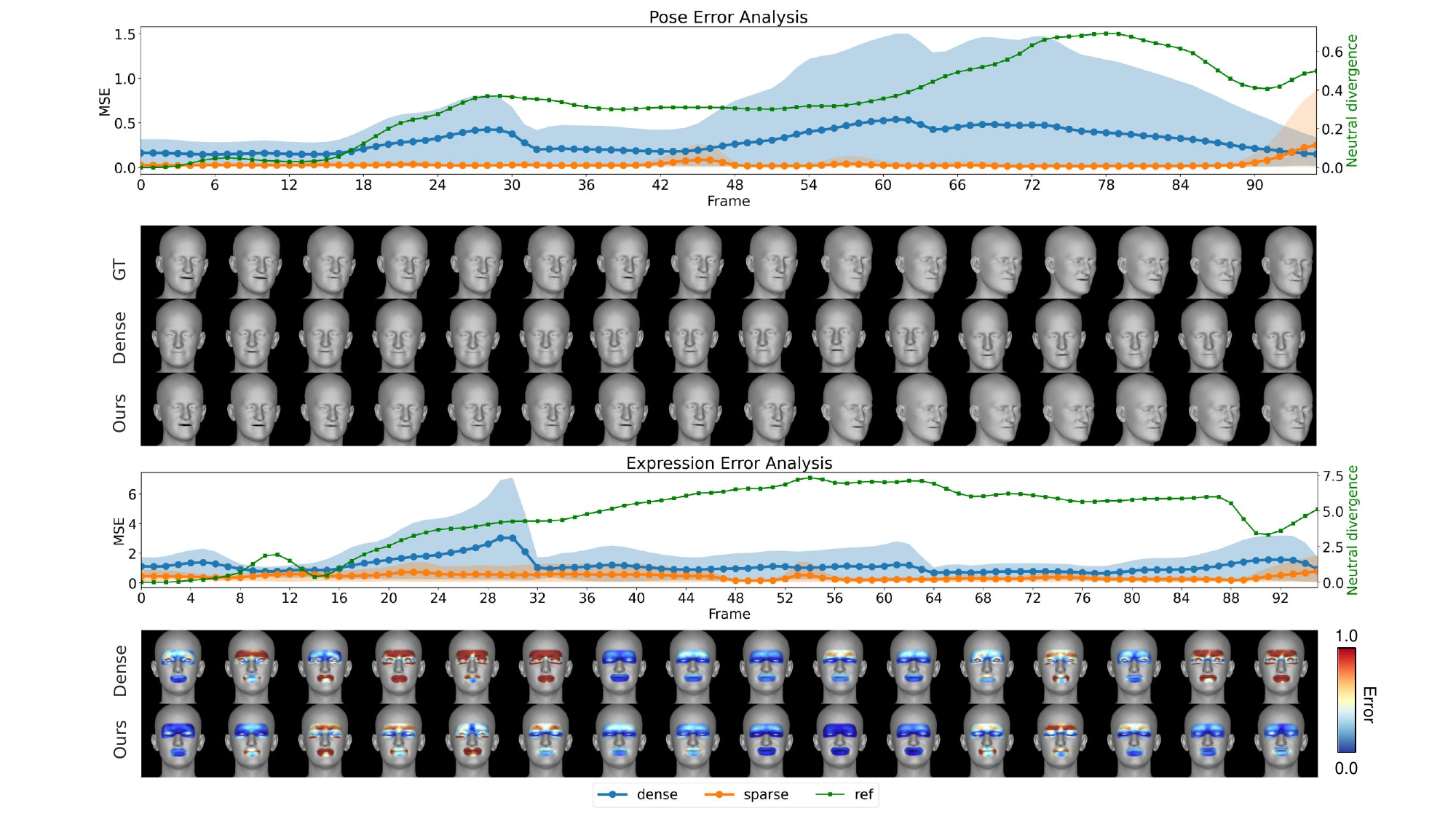}
\caption{\label{fig:error_visual}\textbf{Qualitative comparison on the facial motion reconstruction}. To demonstrate the pose and expression reconstruction accuracy on new motion sequences, we use a DTW-based clustered group analysis. We measure reconstruction error in the same motion cluster and illustrate each MSE loss with a mean error line and variance-shaded area. This visualizes the reconstruction consistency comparison between dense (blue) and sparse (orange) motion representations. The green reference line indicates the distance to a neutral expression, with drastic changes signaling transitions to different sub-motions, challenging the reconstruction. \textbf{Top} For pose control comparison, we measure the MSE of reconstructed poses on a new facial motion sequence to assess the generalization ability of dense and sparse tokenizers. \textbf{Bottom}--- Expression-wise error is visualized via heatmap on two major facial areas: eyes and mouth. Vertice-wise normalized errors for respective areas are colored where colder hues represent lower errors.}
\end{figure*}

Finally, for the second hypothesis on the listening head prediction task, we employed multiple metrics from previous studies in a comprehensive benchmark inherited from two predecessors: L2L \cite{ng2022learning} and REACT23\cite{song2023react2023}. Each is supported and followed by an independent line of work \cite{song2023emotional, liang2023unifarn, dam2024finite} that pursues a different set of metrics for appropriateness, diversity, and synchrony of the generated motion. For comparison, we included the most common metrics in each group.

\textbf{L2L} \cite{ng2022learning} emphasizes subject-wise facial reaction re-creation with:
\begin{enumerate}
    \item L2: distance to corresponding observation motion
    \item Frechet distance (FD): distance between the generated and the ground-truth distribution.
    \item Shannon Index: we run k-means ($K\in\{15,9\}$---an optimal value found by the elbow technique) and compute average entropy of the cluster ID histogram.
    \item Paired FD: distance between the generated and ground-truth concatenated listener-speaker features.
    \item Residual Pearson Correlation Coefficient (RPCC): covariance between the speaker and listener action space: lip curvature and head motion for the expression and the head pose, respectively.
\end{enumerate}

\textbf{REACT23} \cite{song2023react2023}, on the other hand, pays more attention to one-to-many generalization capability of candidate solutions.
\begin{enumerate}
    \item FRDist: the temporally aligned Euclidean distance between the generated and ground-truth facial motion.
    \item FRC: correlation-based metrics to capture the similarity between listener and speaker sequential facial motions.
    \begin{equation}
        \begin{aligned}
            \text{FRC}_{(X,Y)} &= CCC(X,Y) \\
            &= \frac{2\rho\sigma_X\sigma_Y}{\sigma_X^2 + \sigma_Y^2 + (\mu_X-\mu_Y)^2}
        \end{aligned}
    \end{equation}
    \item FRDiv and FRDvs: verify if the model can synthesize diverse motion given the same context. Given $K\times N$ generated motions length $T$, each $N$ context corresponds to $K$ predictions:
    \begin{equation}
        \begin{aligned}
            \text{FRDiv}_{(X)} &= \frac{1}{N}\sum_{i=0}^{N}{L2(x_i,X)^2} \\
            \text{FRDvs}_{(X)} &= \frac{1}{K}\sum_{j=0}^{K}{L2(x_j, X)^2}
        \end{aligned}
    \end{equation}
    \item FRVar: motion variance across the time dimension.
    \begin{equation}
        \text{FRVar}_X = \frac{1}{K\times N}\sum{\bigg(\frac{\sum{(x_t - \hat{x})}}{T-1}\bigg)}
    \end{equation}
    \item FRReal: Frechet Inception Distance (FID) measures the distribution distance between generated facial reaction and ground-truth motions.
    \item FRSync: Time Lagged Cross Correlation (TLCC) verifies the synchrony between the listener and speaker.
\end{enumerate}

According to the performance reported for benchmarks \ref{tab:pred_result1} and \ref{tab:pred_result2}, \textbf{SFMS} with its sparse structure improved the overall quality of the generated listening head facial motion with a smaller network (see Table \ref{tab:recon_result}).

\subsection{Qualitative Comparison}
\subsubsection{Subjective evaluation}
{Given the subjective nature of human perception in non-verbal facial behavior, quantitative metrics alone may not fully capture the expressiveness or appropriateness of generated facial expressions. To address this limitation, we conducted a subjective evaluation comparing the predicted listening head motions of \textbf{SFMS} with two baselines \cite{ng2022learning, dam2024finite}. The evaluation focused on two key criteria: appropriateness and diversity of the generated facial reactions. A total of 25 participants, all university students aged between 23 and 27, took part in the study.}

{In each session, participants were presented with a randomly sampled prediction from \textbf{SFMS} and one of the baselines, and asked to rate which one is more appropriate or expressive using a 5-point comparative rating scale: +2 if one model was significantly better, +1 if slightly better, and 0 if no perceptual difference was observed. For the appropriateness evaluation, participants were shown a single listening head prediction from each model, alongside the corresponding speaker and ground-truth listener video as reference. For the diversity evaluation, a batch of three facial reactions generated by each model was presented without reference, and participants were asked to judge which model exhibited greater variation while remaining contextually plausible.}

\begin{figure}[ht!]
\centering
\includegraphics[width=0.5\textwidth, trim={2.5cm 4.5cm 2cm 3cm}, clip]{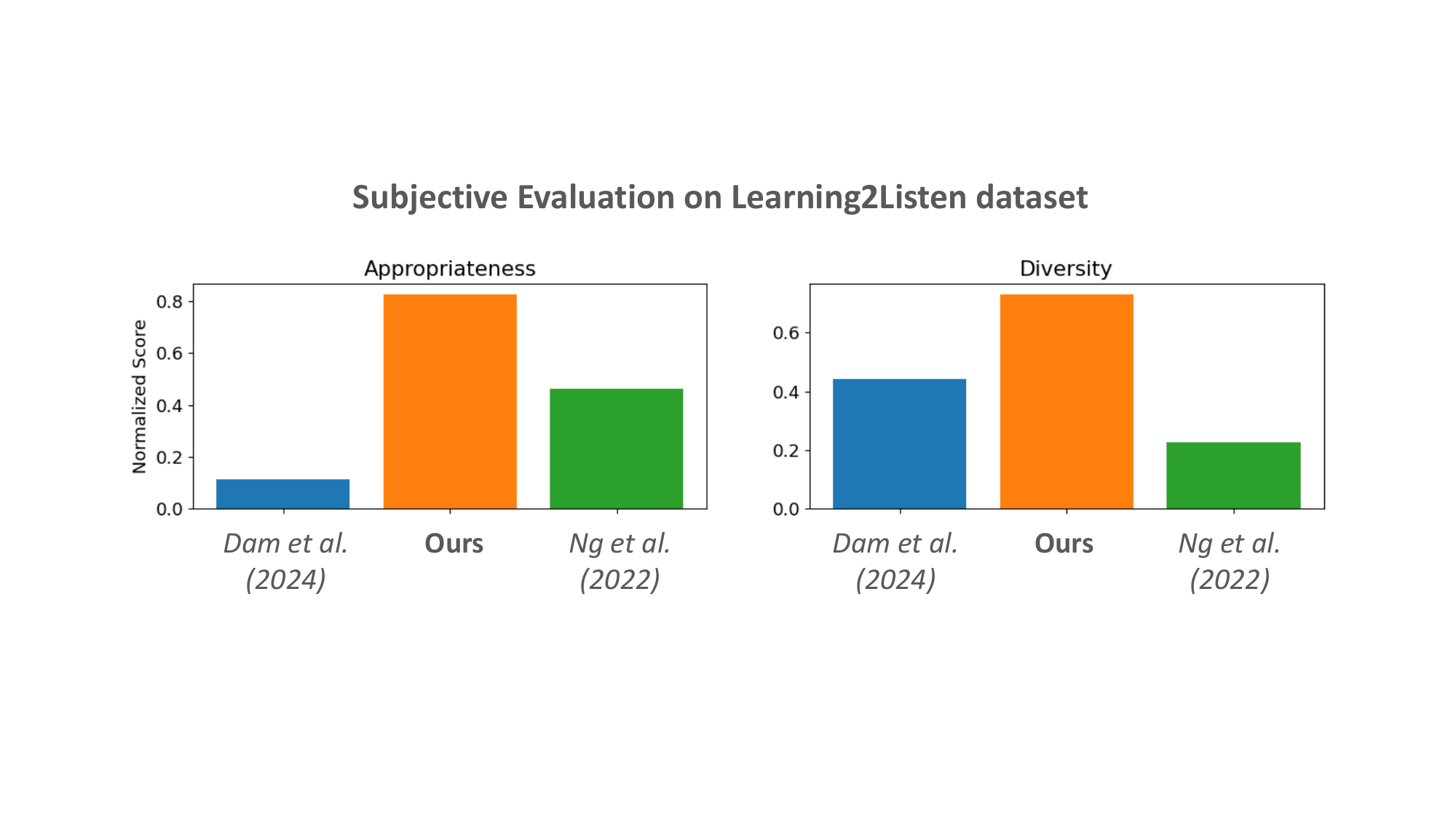}
\caption{\label{fig:subjective_eval} \textbf{Subjective evaluation on L2L dataset}. A pair-wise evaluation was conducted between randomly sampled listening head predictions generated by three candidates. The 5-point comparative rating scale was employed (+2 if model A is significantly better; + 1 if slightly better; 0 if no perceptual difference). Score is normalized by the occurrence count by each model.}
\end{figure}

{As shown in Figure \ref{fig:subjective_eval}, \textbf{SFMS} consistently generated facial reactions that were rated as more appropriate and more diverse compared to those of the competing methods. Comparative video demos are available\footnote{Project page: \url{https://nguyenntt97.github.io/projects/sfms_25}}}

\subsubsection{Facial motion temporal and cluster structure}
Capturing and describing fluid and subtle continuous facial motions are critical for transferring natural nonverbal facial behavior to an agent. Token prediction combining distant patterns is a well-reported issue, owing to the nature of the discrete codebook \cite{ng2022learning, song2023emotional}. This issue is reflected in our observations after a dense tokenization process, in which distinct motion clusters were combined into a larger group. This helps the codebook generalize a wider range of motion, but also perplexes the predictor, which learns to maximize the next token ground truth. To visualize the inspection, we plotted several pairs of contrastive motion clusters (solid and dashed lines), each consisting of several motion members, before and after the transformation, as shown in {Figure \ref{fig:noise_pattern}}. For every pair, we selected a more distinctive set of two clusters (dashed light colored and filled darker colored), where sequence-wise inter- and intra-distances are more discernible.

Both approaches noticeably reduced the temporal variance, as expected from the continuous-to-discrete transformation. Contrary to dense representation, where reconstructed correspondences are shifted toward each other, losing their intensity and temporal characteristics, the sparse counterpart maintains a more appropriate temporal structure and the cluster boundaries are still well defined. This provides a reasonable explanation for the improved performance of the reconstruction and quantitative prediction evaluation.

\subsubsection{Generated facial motion quality} {In Figure \ref{fig:error_visual}}, we visualize the normalized error on mesh vertices between dense and sparse motion structures on two components: eyes and mouth area. This normalization was based on extreme expressions in the dataset, with errors shown as a heatmap mask. A 3D mesh was used for visual inspection of the head pose. The visualized target was a reconstruction of the same cluster member motion. In addition, the chart displays the average error for all the instances within the cluster. According to our experimental results, the proposed method reconstructed the motion more consistently, with a noticeable improvement in accuracy.

\section{Conclusion and Future Work}

We propose \textbf{SFMS}, a sparse structure designed to capture the temporal continuous dynamics of 3DMM-based facial nonverbal features from video datasets. Our method leverages keyframe elements in an unsupervised manner to learn a finite facial motion codebook for given subjects, and successfully applies this to future listening reaction prediction tasks. Experimental results demonstrate that our model significantly improves both quantitative and qualitative performance in nonverbal facial motion representation and listening head prediction.

However, several limitations remain: first, employing a dynamic keyframe number strategy could provide further improvements depending on some situations.  Secondly, the two public datasets, {despite being the bigger ones for the listening head prediction task}, are significantly smaller compared to their counterparts in the talking head generation task. {Verifying how scaling affects sparse representation similar to \textbf{SFMS} is interesting to further improve facial motion-related tasks. Finally, despite not being tested in this study, the proposed sparse representation provides a unique domain-specific attention score, aligning with recent demand for longer context and more efficient training \cite{yuan2025native}.}

\bibliographystyle{ieeetr}
\bibliography{refs}

\vspace{11pt}




\vfill

\end{document}